\documentclass[letterpaper]{IEEEtran}
\usepackage{amsmath,amssymb,amsfonts}
\usepackage{algorithmic}
\usepackage{graphicx}
\usepackage{textcomp}
\usepackage{color}
\usepackage{xcolor}
\usepackage{colortbl}
\usepackage{subcaption}
\usepackage{url}
\usepackage[ruled,linesnumbered]{algorithm2e}
\usepackage{array}
\usepackage{multirow}
\usepackage{booktabs}

\begin{document}

\title{Whole-body Motion Control of an Omnidirectional Wheel-Legged Mobile Manipulator via Contact-Aware Dynamic Optimization}

\author{Zong Chen, Shaoyang Li, Ben Liu, Min Li, Zhouping Yin and Yiqun Li* 

\thanks{This work was supported in part by the National Natural Science Foundation of China No. 51905185, the Interdisciplinary Research Support Program of HUST No. 2024JCYJ040, and the National Postdoctoral Program for Innovative Talents No. BX20180109.

Zong Chen, Shaoyang Li, Ben Liu, Min Li, Zhouping Yin and Yiqun Li are with the State Key Laboratory of Intelligent Manufacturing Equipment and Technology, School of Mechanical Science and Engineering, Huazhong University of Science and Technology, Wuhan, 430074, China. (e-mail: {\tt\small skelon\_chan@hust.edu.cn; liyiqun@hust.edu.cn}). 
}%
}


\maketitle

\begin{abstract}
Wheel-legged robots with integrated manipulators hold great promise for mobile manipulation in logistics, industrial automation, and human-robot collaboration. However, unified control of such systems remains challenging due to the redundancy in degrees of freedom, complex wheel-ground contact dynamics, and the need for seamless coordination between locomotion and manipulation. In this work, we present the design and whole-body motion control of an omnidirectional wheel-legged quadrupedal robot equipped with a dexterous manipulator. The proposed platform incorporates independently actuated steering modules and hub-driven wheels, enabling agile omnidirectional locomotion with high maneuverability in structured environments. To address the challenges of contact-rich interaction, we develop a contact-aware whole-body dynamic optimization framework that integrates point-contact modeling for manipulation with line-contact modeling for wheel-ground interactions. A warm-start strategy is introduced to accelerate online optimization, ensuring real-time feasibility for high-dimensional control. Furthermore, a unified kinematic model tailored for the robot's 4WIS-4WID actuation scheme eliminates the need for mode switching across different locomotion strategies, improving control consistency and robustness. Simulation and experimental results validate the effectiveness of the proposed framework, demonstrating agile terrain traversal, high-speed omnidirectional mobility, and precise manipulation under diverse scenarios, underscoring the system's potential for factory automation, urban logistics, and service robotics in semi-structured environments.
\end{abstract}


\IEEEpeerreviewmaketitle

\section{Introduction}

\begin{figure}[t]
    \centering
    \includegraphics[width=0.45\textwidth]{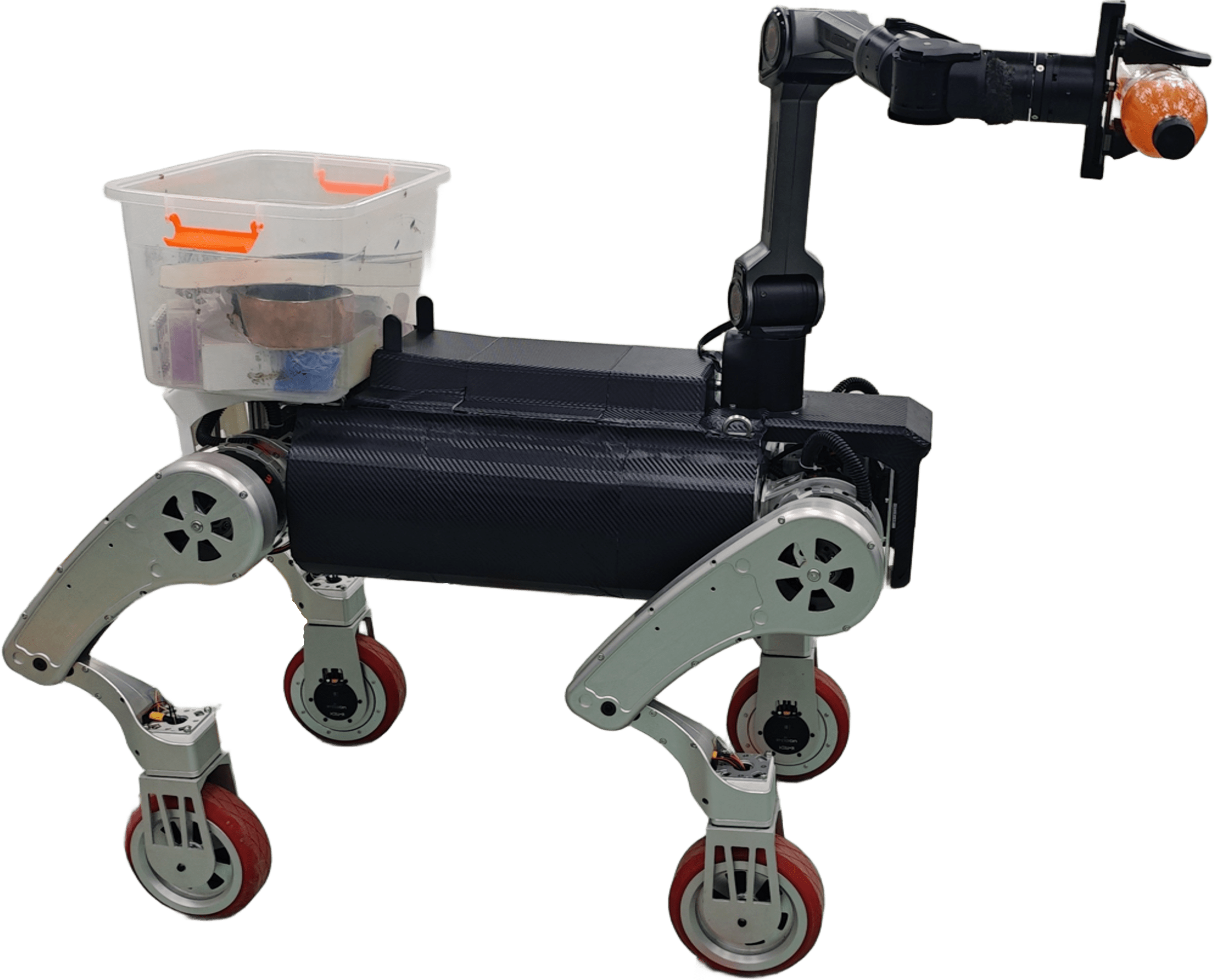}
    \caption{An Omnidirectional Wheel-Legged Robot with a Dexterous Manipulator}
    \label{fig_robo}
\end{figure}

\IEEEPARstart{W}{heel-legged} quadrupedal robots have demonstrated strong potential for applications such as material transport in factories and indoor-outdoor logistics, owing to their superior mobility and terrain adaptability. By integrating a dexterous manipulator, such robots are further equipped with autonomous manipulation capabilities, enabling tasks such as grasping, handling, sorting, and delivery (see Fig.~\ref{fig_robo}). Recently, wheeled, legged, and hybrid wheel-legged systems have been increasingly deployed in industrial automation, search and rescue, and logistics. However, both wheeled and legged robots exhibit inherent limitations in terms of mobility and manipulation: wheeled robots struggle to traverse uneven terrain, while legged robots often suffer from low efficiency, high energy consumption, and limited speed when operating on flat surfaces. In practical scenarios, adapting to diverse environments often necessitates switching between distinct platforms, leading to increased operational complexity and maintenance costs.
\par
To address these challenges, wheel-legged robots have emerged as a promising hybrid mobility solution that combines the strengths of wheeled and legged locomotion. These systems offer enhanced dynamic stability and superior adaptability across varied terrains, and have become a focal point of current research \cite{bjelonic2019keep,li2024safe,wang2024arm}. A typical wheel-legged quadruped features 12 actuated joints and 4 driven wheels, offering general maneuverability and versatility. However, structural limitations still hinder their agility on paved surfaces, making it difficult to achieve high-performance maneuvers such as rapid in-place rotations and quiet directional changes.

\par
\par
The integration of manipulators into mobile robots significantly broadens their range of executable tasks. However, this also introduces substantial control challenges, such as increased redundancy in degrees of freedom, uncertainty of the center of mass (CoM) position, and heightened system-level dynamic complexity. In recent years, several research groups have explored mobile robotic systems equipped with manipulators. Representative platforms include Boston Dynamics' Spot robot \cite{hoffman2025learning,zimmermann2021go}, ETH Zurich's ANYmal platform \cite{bjelonic2019keep, bjelonic2020rolling}, and Jueying X20 \cite{deeprobotics2022jueying}. These systems have demonstrated notable progress in coordinated mobility and manipulation. However, most existing studies focus on the coordination between legged bases and manipulators, typically assuming point contacts at the feet. This assumption simplifies the contact modeling process using friction cone constraints. In contrast, wheeled systems present more complex wheel-ground interactions, which require accounting for both rolling constraints and line-contact behaviors, thereby substantially complicating whole-body control modeling. Recently, reinforcement learning (RL) methods have been increasingly applied to whole-body control, demonstrating high performance and robustness \cite{fu2023deep,wang2024arm,ma2025learning}. Nonetheless, RL-based approaches still face substantial challenges in policy generalization across different robotic platforms and achieving high-precision manipulation.
\par
Substantial advancements have been made in both locomotion and manipulation for legged and wheel-legged robots. Based on their underlying control paradigm, existing approaches can be broadly categorized into two types: model-based control and data-driven learning-based control. Model-based approaches rely on precise system modeling and solve optimal control problems based on dynamic equations. Such approaches often employ Model Predictive Control (MPC) to optimize foot contact forces, and Whole-Body Control (WBC) to coordinate joint-level actions for multi-task execution. On the other hand, learning-based methods, such as reinforcement learning (RL) leverage GPU-accelerated computation and deep neural networks to efficiently train control policies. However, the gap between simulation and real-world deployment (the sim-to-real gap) limits the real-world reliability of such methods, particularly when transferring policies across robots with differing morphologies, where generalization and precise task execution remain significant challenges.
\par
For quadruped robots, both model-based and learning-based control strategies have demonstrated robust performance and effective terrain traversal in complex and rugged environments \cite{kolter2008control, lee2020learning, shi2023terrain}. However, the integration of a manipulator significantly increases the complexity of controlling the CoM trajectory, and the challenges of whole-body coordination, high degrees of redundancy, and increased control dimensionality make the overall control problem substantially more difficult. Zimmermann et al. \cite{zimmermann2021go} explored dynamic grasping tasks using the Boston Dynamics Spot robot, but due to restricted access to the low-level controller, their system experienced delays or failures when handling high-disturbance operations. In \cite{ma2022combining}, a hybrid control strategy was proposed that combines Model Predictive Control and reinforcement learning to control the base and manipulator separately, enabling coordinated control. However, such methods generally treat the manipulator as a disturbance to the locomotion controller rather than addressing whole-body motion planning and control as a unified optimization problem. In contrast, several studies \cite{bellicoso2019alma, sleiman2021unified, zhang2024whole, aguirre2024development} have proposed unified control frameworks that tightly integrate the control of the mobile base and manipulator. These approaches formulate full-body dynamics models and adopt receding horizon optimization strategies to enable dynamic execution of complex tasks such as door opening and human-robot collaborative manipulation.
\par
Moreover, with the evolution of whole-body reinforcement learning techniques, robotic systems have shown enhanced adaptability in both motion control and interaction with diverse environments. Existing research has demonstrated that quadruped robots trained with whole-body policies can robustly traverse challenging terrains \cite{lee2020learning, jenelten2024dtc, hoeller2024anymal}, while also performing manipulation tasks with considerable adaptability \cite{yokoyama2023asc, arcari2023bayesian, hoffman2025learning}. For example, Fu et al. \cite{pmlr-v205-fu23a} proposed a unified policy for coordinating quadrupedal locomotion and manipulation across different operation modes. However, their method still requires manual intervention during task mode switching, which may introduce discontinuities or instability during execution. To further improve motion efficiency, integrating wheeled mechanisms into legged robots enables faster and smoother locomotion on flat terrain. Hybrid wheel-legged systems have attracted increasing attention due to their unique dynamic characteristics and structural flexibility \cite{lee2024learning, jelavic2021combined, bjelonic2020rolling}. However, such systems present more complex contact dynamics, imposing higher demands on optimization speed and solver efficiency. To enhance real-time performance, simplified wheel-ground contact models have been proposed \cite{bjelonic2021whole, medeiros2020trajectory}, approximating line contact as point contact subject to friction cone constraints, thereby sacrificing some modeling accuracy in exchange for computational efficiency and numerical stability. In \cite{jiang2024learning}, a novel reward fusion mechanism was introduced to nonlinearly integrate multiple task objectives, enabling whole-body control of wheel-legged robots across various manipulation tasks. Nevertheless, the platform lacks dedicated steering actuators, which results in mechanical oscillations and discontinuous trajectories during large-angle or continuous turning tasks, thereby limiting its applicability in precision manipulation scenarios.
\par
In addition, recent studies have further explored alternative wheel-legged configurations, such as bipedal wheel-legged systems \cite{wang2024arm,yu2023modeling,boston2017handle} and multi-wheel-legged systems \cite{rehman2016towards,dadiotis2022trajectory}. The former, which combines dual wheels with leg-like structures, is prone to significant base disturbances during complex manipulation tasks, limiting its applicability in high-precision or heavy-load operations. The latter, despite providing superior terrain adaptability and dynamic stability due to its multiple wheel supports, suffers from excessive an excessive number of joints, high control redundancy, and system complexity, resulting in reduced locomotion efficiency. These characteristics typically constrain their use to highly unstructured or extreme terrain scenarios.
\par
Despite substantial progress in the coordinated control of wheel-legged robots equipped with manipulators, achieving unified whole-body motion control under wheel-ground frictional constraints remains a major open challenge. In particular, for high-degree-of-freedom wheel-legged platforms equipped with independent steering actuators, the redundancy in motion modes complicates the control architecture, and a unified and efficient solution is still lacking. To address these challenges, we design and develop an omnidirectional wheel-legged robotic platform equipped with eight leg joint actuators, four independent steering motors, and four hub-driven wheels, along with a dexterous manipulator arm (see Fig.~\ref{fig_robo}). This platform combines strong terrain traversal capability with agile omnidirectional locomotion on flat surfaces, enabling both high-speed mobility and precise manipulation tasks. Building upon this platform, we further propose a whole-body motion control framework grounded in contact dynamics. A unified multibody optimization model is constructed, integrating point-contact modeling for the manipulator and line-contact modeling for wheel-ground interactions. By using the results of low-frequency motion planning as $warm-start$ inputs, the proposed framework supports high-frequency real-time optimization and control. In addition, a unified kinematic model tailored for the platform's specific wheel actuation mechanism is developed, which obviates the need for switching control strategies based on different locomotion modes, thereby simplifying the control logic and improving execution efficiency and system robustness. The main contributions of this study are summarized as follows,
\begin{itemize}
  \item Design and implementation of an omnidirectional manipulation-capable wheel-legged robot: Compared with traditional 12-DOF legged robots with 4 wheels, the proposed platform incorporates independent steering actuators at each leg end, significantly enhancing omnidirectional mobility on flat terrain and improving the flexibility of the onboard manipulator.
  \item A contact-dynamics-based whole-body motion planning and control framework: A unified contact model is developed to simultaneously account for point contacts of the manipulator and line contacts of the wheels. An online $warm-start$ optimization strategy is used to realize real-time control of high-DOF robotic systems.
  \item A unified wheeled-motion kinematic optimization model: This model avoids the frequent switching of control frameworks under different motion modes, thereby improving consistency, real-time capability, and overall control efficiency for omnidirectional wheel-legged robots.
\end{itemize}

\section*{Methods}
\subsection*{Omnidirectional Wheel-legged Robot with Manipulator}
\begin{figure}[htbp]
    \centering
    \includegraphics[width=0.48\textwidth]{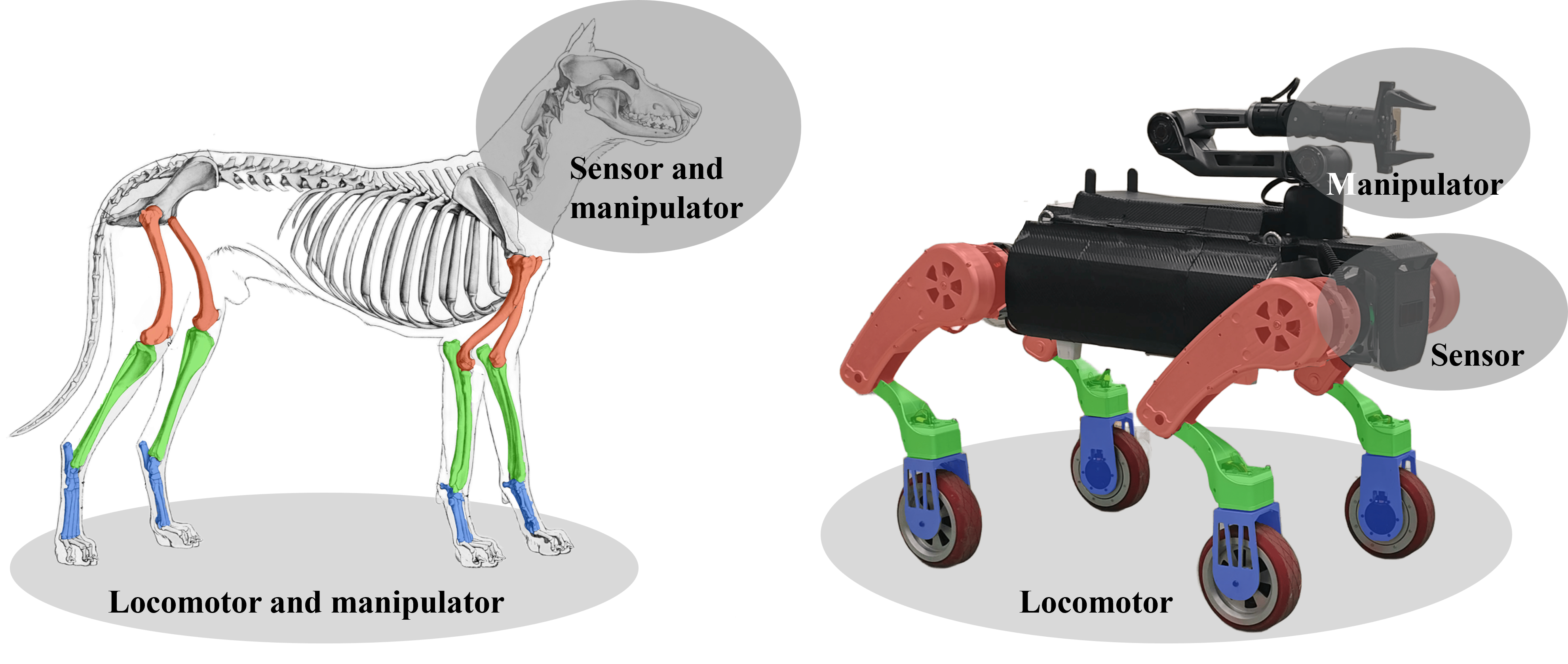}
    \caption{Biologically inspired comparison between a quadruped animal (left) and the proposed omnidirectional wheel-legged robot (right). }
    \label{fig_robot_structure}
\end{figure}

Quadrupedal robots are widely adopted in complex environments due to their superior terrain traversal capabilities. To enhance locomotion efficiency on flat and paved surfaces, active wheel modules are often integrated at the end of each leg, resulting in the widely adopted wheel-legged configuration. Expanding upon this concept, we introduce independently actuated steering modules to significantly improve the robot's omnidirectional mobility. Based on our prior work \cite{li2024simulation, li2024safe}, we design and implement a biomimetic omnidirectional wheel-legged robotic system equipped with a manipulator, as illustrated in Fig.~\ref{fig_robot_structure}. Compared to conventional wheel-legged platforms with hip-actuated joints, this platform sacrifices a degree of dynamic stability on rough terrain but achieves markedly improved mobility and manipulation flexibility in urban streets, factory floors, and other structured environments. The structural design seeks to balance terrain adaptability with operational efficiency, making it well-suited for mobile manipulation tasks requiring high agility.
\par
To emulate the locomotion, sensing, and manipulation capabilities of quadrupedal animals, as shown in Fig.~\ref{fig_robot_structure}, the proposed platform adopts a biologically inspired design. Each leg is equipped with a 360-degree independently actuated steering module, which, together with hub-driven wheels, enables fully omnidirectional motion—such as in-place rotation (zero turning radius), crab walking, and other agile maneuvers. Coupled with a 7-DOF manipulator (including an end-effector gripper), the platform is capable of performing complex locomotion and manipulation tasks even in confined environments. Additionally, the robot integrates multiple perception sensors, including depth cameras and LiDAR, to support environment perception and map construction. Together with onboard autonomous navigation algorithms, the system enables global path planning and fully autonomous motion control, providing strong perception and decision-making capabilities for manipulation tasks in both indoor and outdoor environments.

\subsection*{Dynamic Model with Contact}
\subsubsection*{Whole-Body Multibody System Modeling}
\begin{figure}[t]
    \centering
    \includegraphics[width=0.45\textwidth]{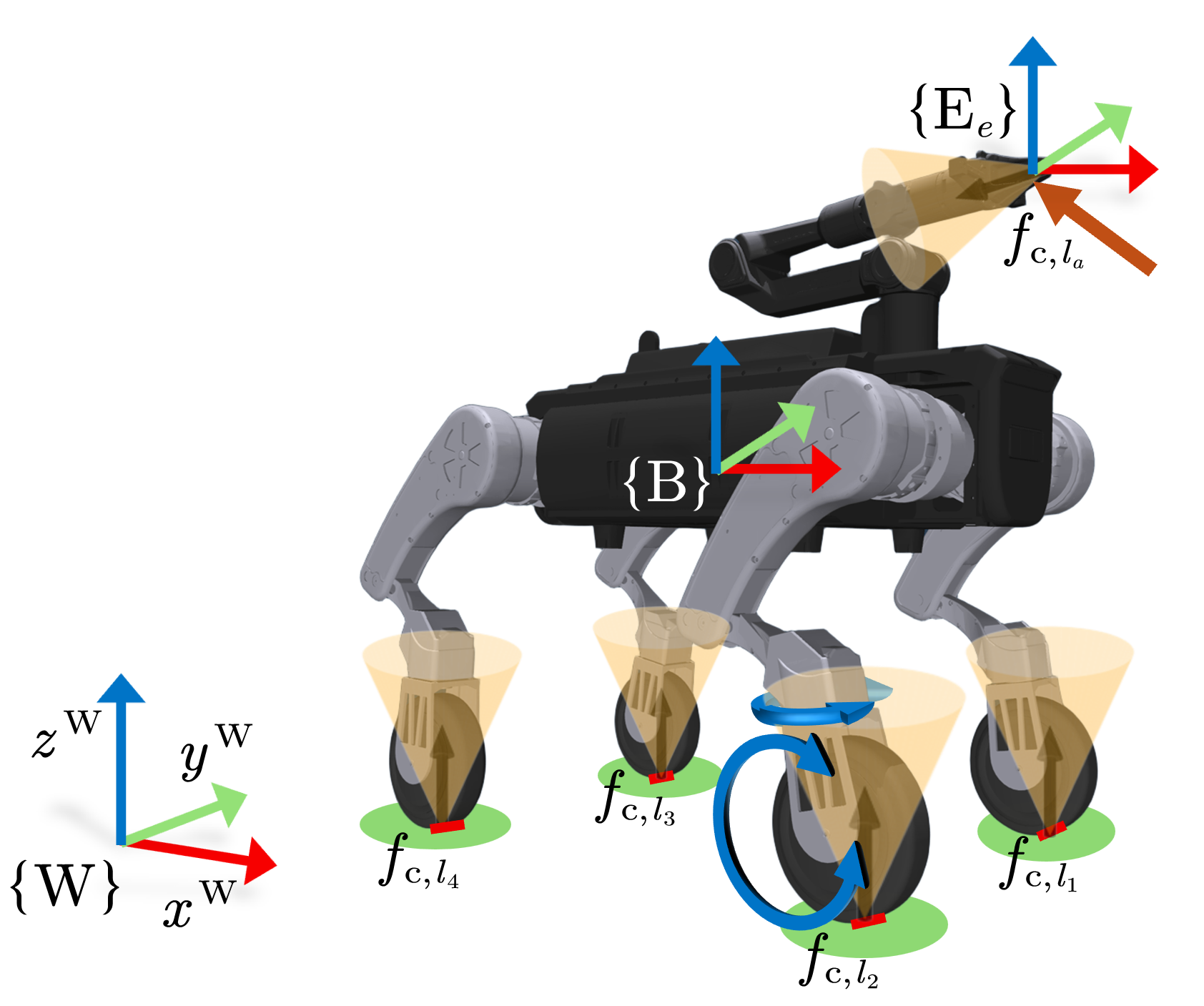}
    \caption{Frames and contact model: ${\{\mathrm B\}}$, end-effector ${\{\mathrm {E}_e\}}$, and ${\mathrm{c}_i}$; friction cones (yellow), contact line (red), and a green ellipse jointly defining the degenerate 6-D wrench constraint.}
    \label{fig_robot_model}
\end{figure}

To facilitate unified control of both the mobile base and the manipulator, we extend the conventional multibody modeling approach into a whole-body dynamics framework encompassing the entire robot system. In contrast to traditional approaches that model the mobile platform and the manipulator independently, our approach treats the combined system as a single multibody structure, making it more suitable for unified planning and control in whole-body optimization tasks. Similar to standard legged robot models, the limbs equipped with driving wheels and steering actuators are modeled as leg modules, while the manipulator and end-effector gripper are treated as a single end-effector unit, all connected to a floating base, as illustrated in Fig.~\ref{fig_robot_model}. In the fixed world frame $\{\mathrm{W}\}$, the robot's generalized position $q$ and generalized velocity $\dot{q}$ are defined as,
\begin{equation}
    q\,\,=\,\,\left[ \begin{array}{c}	r_{\mathrm{WB}}\\	q_{\mathrm{WB}}\\	q_l\\	q_a\\\end{array} \right] \in SE\left( 3 \right) \times \mathbb{R} ^{n_l+n_a},\,\,\, \,\,\, \dot{q}=\left[ \begin{array}{c}	v_{\mathrm{WB}}\\	\omega _{\mathrm{WB}}\\	\dot{q}_l\\	\dot{q}_a\\\end{array} \right] \in \mathbb{R} ^{n_v}
    \label{eq_robot_pose}
\end{equation}
here, $r_{\mathrm{WB}} \in \mathbb{R}^3$ represents the position of the robot's center of mass in the world frame; $q_{\mathrm{WB}} \in SO(3)$ represents the base orientation; $q_l \in \mathbb{R}^{n_l}$ and $q_a \in \mathbb{R}^{n_a}$ denote the joint angles of the legs and the manipulator, respectively. Similarly, $v_{\mathrm{WB}} \in \mathbb{R}^3$ and $\omega_{\mathrm{WB}} \in \mathbb{R}^3$ are the linear and angular velocities of the base; $\dot{q}_l \in \mathbb{R}^{n_l}$ and $\dot{q}_a \in \mathbb{R}^{n_a}$ are the angular velocities of the legs and the manipulator joints. The total dimension of the generalized velocity $q$ is $n_v = 6 + n_l + n_a$.
Based on these definitions, the dynamics of the whole locomotion-manipulation system can be uniformly expressed as,
\begin{equation}
    M(q) \ddot{q} + h(q, \dot{q}) = \mathcal{S} ^\text{T} u + J_c(q)^\text{T} f_{\text{c}}
    \label{eq_robot_dynamics}
\end{equation}
where $M(q) \in \mathbb{R}^{n_v \times n_v}$ is the generalized inertia matrix, and $h(q, \dot{q}) \in \mathbb{R}^{n_v}$ includes Coriolis, centrifugal, and gravitational effects. $\mathcal{S}^\text{T} u \in \mathbb{R}^{n_v}$ represents the actuation forces generated by the joints (legs and manipulator), with $\mathcal{S} \in \mathbb{R}^{n_u \times n_v}$ being the selection matrix that omits unactuated degrees of freedom of the floating base. The control input $u \in \mathbb{R}^{n_u}$ typically represents the commanded joint torques. $J_c(q) \in \mathbb{R}^{n_c \times n_v}$ is the Jacobian matrix at the contact points, and $f_{\text{c}} \in \mathbb{R}^{n_c}$ represent the contact force vector.
\par
During locomotion and manipulation tasks, the robot maintains physical contacts with the environment through its foot ends and the manipulator end-effector. To ensure physical consistency and prevent interpenetration, it is necessary to enforce that the acceleration at all contact points remains zero along the contact normal direction. This constraint can be formally expressed as,
\begin{equation}
    J_c(q) \ddot{q} + \dot{J}_c(q, \dot{q}) \dot{q} = 0
    \label{eq_contact_constraint}
\end{equation}
where $J_c(q)$ is the Jacobian matrix at the contact points, and $\dot{J}_c(q, \dot{q})$ denotes its time derivative.
\par
By combining the above contact constraint \eqref{eq_contact_constraint} with the system dynamics equation \eqref{eq_robot_dynamics}, a coupled dynamic-constrained system that incorporates contact reaction forces is obtained as follows,
\begin{equation}
    \begin{bmatrix} M(q) & -J_c^\text{T}(q) \\ J_c(q) & 0 \end{bmatrix} \begin{bmatrix} \ddot{q} \\ f_{\text{c}} \end{bmatrix} = \begin{bmatrix} \mathcal{S} ^\text{T} u - h(q, \dot{q}) \\ -\dot{J_c}(q, \dot{q}) \dot{q} \end{bmatrix}
    \label{eq_robot_dynamics_contact}
\end{equation}
In the absence of contact, this system degenerates into the standard unconstrained dynamics model,
\begin{equation}
    \ddot{q} = M(q)^{-1}(\mathcal{S} ^\text{T} u - h(q, \dot{q}))
    \label{eq_robot_dynamics_no_contact}
\end{equation}
which describes the dynamic behavior of the leg or arm in airborne states.
\par
To ensure non-slipping and non-rolling behavior at contact points during locomotion and manipulation, it is essential to impose physical constraints on the contact forces. In traditional legged robots, the ground contact at the end-effector is often simplified as a point contact, where the contact force is typically modeled using a friction cone. For systems with larger contact surfaces—such as humanoid robots—contact modeling usually adopts a full six-dimensional wrench constraint framework to capture the coupling between forces and moments \cite{crocoddyl20icra, chatzinikolaidis2021trajectory}. In contrast, the wheel-ground contact in wheel-legged robots belongs to the category of line contact. Common approaches include discretized point approximations \cite{drumwright2010extending_ode, coumans2015bullet}, equivalent contact point modeling \cite{xie2016rigid}, and the formulation of hybrid dynamic systems with nonholonomic constraints to capture pure rolling and no-slip conditions \cite{sarkar1994control}. In addition, slip-based friction models from vehicle dynamics \cite{kalker1966rolling, jaswin2020applicationkalkertheoryrolling} have been introduced to simulate the complex wheel-ground interaction. To enable efficient real-time control in this study, we adopt a simplified (degenerated) form of the full wrench contact model to approximate wheel-ground interactions. As illustrated in Fig.~\ref{fig_robot_model}, a three-dimensional friction cone is applied to the contact force, along with constraints on the normal and lateral torques at the contact interface. The torque along the rolling-direction is left unconstrained to reduce the modeling complexity. For the manipulator's point contact at the end-effector, conventional friction cone constraints are still applied.
\par
For each contact, we define the local contact wrench as,
$$ \boldsymbol f_{\mathrm c} = [f_x,f_y,f_z,\tau_x,\tau_y,\tau_z]^\mathrm{T} \in \mathbb{R}^6 $$
\par
To enhance optimization tractability, the nonlinear Coulomb friction cone is approximated by a polyhedral (linear) contact-wrench cone (CWC), enabling us to cast contact constraints into a standard H-representation (i.e., a set of half-space inequalities):
\begin{equation}
    \mathbf{A}_{\mathrm c} \,\boldsymbol f_{\mathrm c} \le \mathbf{b}_{\mathrm c}
    \label{eq_friction_cone}
\end{equation}
where $\mathbf{A}_\text{c} \in \mathbb{R}^{m \times 6}$ is the face-normal (inequality) representation of the contact-wrench cone, and $\mathbf{b}_\text{c} \in \mathbb{R}^m$ is the upper-bound vector.
\par
For standard full six-dimensional contact modeling, friction cone and torque limits can be represented as linear constraints in the form of a dual \{$\mathbf{A}_\text{c}$, $\mathbf{b}_\text{c}$\},
\begin{equation}
\mathbf{A}_\text{c} = \left[
\begin{array}{rrrrrr}
\mu & 0 & -1 & 0 & 0 & 0 \\ 
-\mu & 0 & -1 & 0 & 0 & 0 \\
0 & \mu & -1 & 0 & 0 & 0 \\
0 & -\mu & -1 & 0 & 0 & 0 \\
0 & 0 & 0 & 1 & 0 & 0 \\
0 & 0 & 0 & -1 & 0 & 0 \\
\rowcolor{gray!15} 0 & 0 & 0 & 0 & 1 & 0 \\ 
\rowcolor{gray!15} 0 & 0 & 0 & 0 & -1 & 0 \\ 
0 & 0 & 0 & 0 & 0 & 1 \\ 
0 & 0 & 0 & 0 & 0 & -1
\end{array}
\right], \,
\mathbf{b}_\text{c} = \begin{bmatrix}
0 \\ 0 \\ 0 \\ 0 \\
\tau_{x, \text{max}} \\ \tau_{x, \text{min}} \\
\tau_{y, \text{max}} \\ \tau_{y, \text{min}} \\
\tau_{z, \text{max}} \\ \tau_{z, \text{min}}
\end{bmatrix}
\label{eq_friction_cone_full}
\end{equation}
The first four rows of $\mathbf{A}_\text{c}$ approximate the Coulomb friction cone: $|f_x| \leq \mu f_z$, $|f_y| \leq \mu f_z$. The remaining six rows bound the angular components (moments) of the contact wrench about the $x$, $y$, and $z$ axes, where $\tau_{*,\{max/\min\}}$ denote the allowable limits in each direction.
\par
For point contact modeling of the manipulator end-effector, only the first four friction cone constraints are retained. An additional non-negative normal force constraint $f_z \geq 0$ is typically enforced to ensure the physical consistency of contact direction. In the case of wheel-ground line contact, we exploit the contact structure to reduce complexity. Specifically, we use a degenerate contact-wrench model: the friction cone in ($f_x$, $f_y$, $f_z$) is kept, and moment bounds about the $x$ (lateral) and $z$ (normal) axes are enforced, while the moment about the $y$ axis (i.e., the rolling direction) is left unconstrained in our inequality set. Accordingly, rows 7-8 of $\mathbf{A}_\text{c}$ are set to be zero, removing the corresponding moment limits; this yields a compact, physically reasonable approximation of wheel-ground interaction.
\par
In constrained optimization problems, hard constraints often pose challenges to convergence and feasibility, especially in complex systems where they may lead to infeasibility or numerical instability \cite{marquez2017imposing}. To address this, the linear inequality constraints arising from contact modeling are transformed into residual-based soft constraints and embedded into the objective function. Specifically, for the inequality constraint $\mathbf{A}_\text{c} \boldsymbol{f_{\text{c}}} \leq \mathbf{b}_\text{c}$, the corresponding residual function is defined as,
\begin{equation}
    \mathbf{r}(\boldsymbol{f_{\text{c}}}) = \mathbf{A}_\text{c} \boldsymbol{f_{\text{c}}} - \mathbf{b}_\text{c}
    \label{eq_friction_cone_residual}
\end{equation}
To penalize violations of the friction cone constraint, we introduce a squared penalty function defined as,
\begin{equation}
    \alpha(\mathbf{r}) =
    \begin{cases}
        0, & \text{if } \mathbf{r} \leq 0 \\
        \frac{1}{2} \mathbf{r}^\mathrm{T} \mathbf{r}, & \text{if } \mathbf{r} > 0
    \end{cases}
    \label{eq_friction_cone_penalty}
\end{equation}
This penalty term $\alpha(\mathbf{r})$ is incorporated into the overall objective function as a soft constraint cost, representing both friction cone and torque limit violations.
\par
In a multibody system, contact wrenchs must simultaneously satisfy the friction cone constraints and the global dynamic equilibrium. Specifically, for the $i$-th contact point, the contact wrench $\boldsymbol{f_{\text{c},i}}$ and contact-wrench constraint matrix $A_{c,i}$ are subject to the following linear inequality,
\begin{equation}
    A_{c,i} \boldsymbol{f_{\text{c},i}} \le 0, \quad \forall i
    \label{eq_friction_cone_i}
\end{equation}
such that the set of admissible global contact wrenches forms the contact-wrench cone. Simultaneously, the system must satisfy the Newton-Euler equilibrium in generalized coordinates,
\begin{equation}
    \sum_i J_i^\text{T} \boldsymbol{f_{\text{c},i}} + \boldsymbol{f}_\text{ext} = 0
    \label{eq_momentum_balance}
\end{equation}
where $\boldsymbol{f}_\text{ext}$ denotes the sum of external generalized forces acting on the system.
\par
Under friction cone constraints and Newton-Euler equilibrium conditions, one can define the Zero Moment Point (ZMP) as the projection of resultant wrench onto a reference plane via $p_{\mathrm{zmp}}=\left( n\times \tau _O \right) /\left( n\cdot f \right) $. Caron et al. \cite{caron2016zmp} showed that the image of the CWC under this projection is the full support area; if contacts are coplanar and friction is sufficiently large, this area reduces to the convex hull of contact points, yielding the classical ZMP stability condition (a necessary condition). Under the Linear Pendulum Mode (LPM)—i.e., added CoM-height constancy and angular-momentum regulation—this support domain further contracts to the pendular support area, and ZMP lying within it becomes a necessary and sufficient condition for multi-contact stability.
\subsubsection*{Unified Model for Omnidirectional Mobile Robots}
\begin{figure*}[h]
    \centering
    \begin{subfigure}[b]{0.24\textwidth}
        \includegraphics[width=0.9\textwidth]{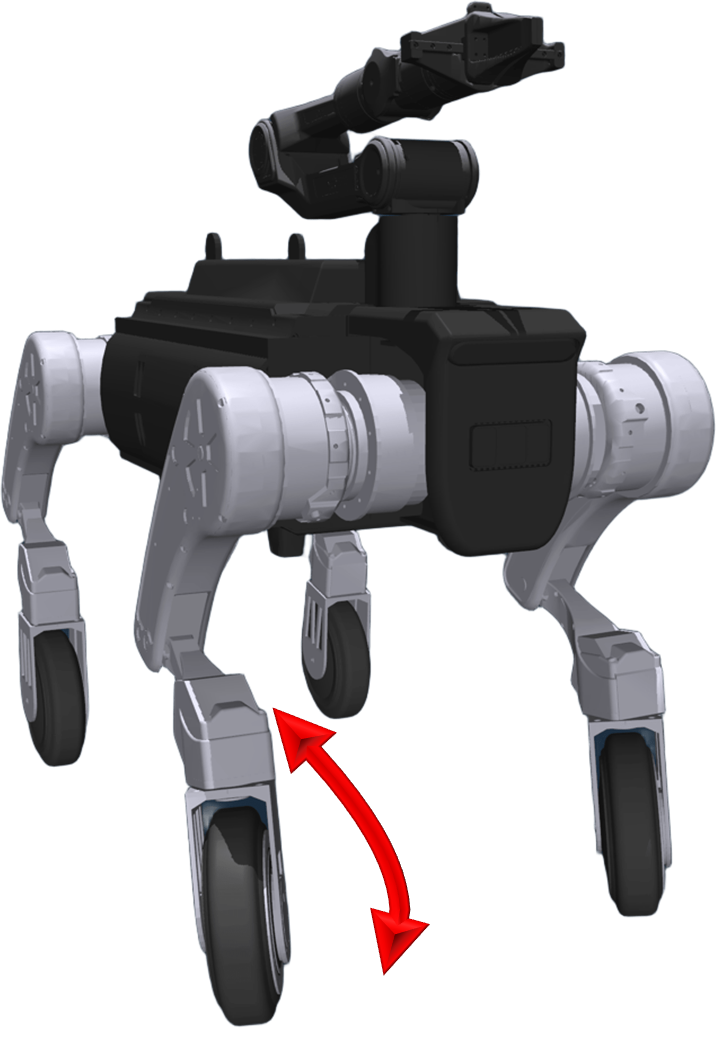}
        \caption{Ackermann steering}
        \label{fig_ackman}
    \end{subfigure}
    \begin{subfigure}[b]{0.24\textwidth}
        \includegraphics[width=0.9\textwidth]{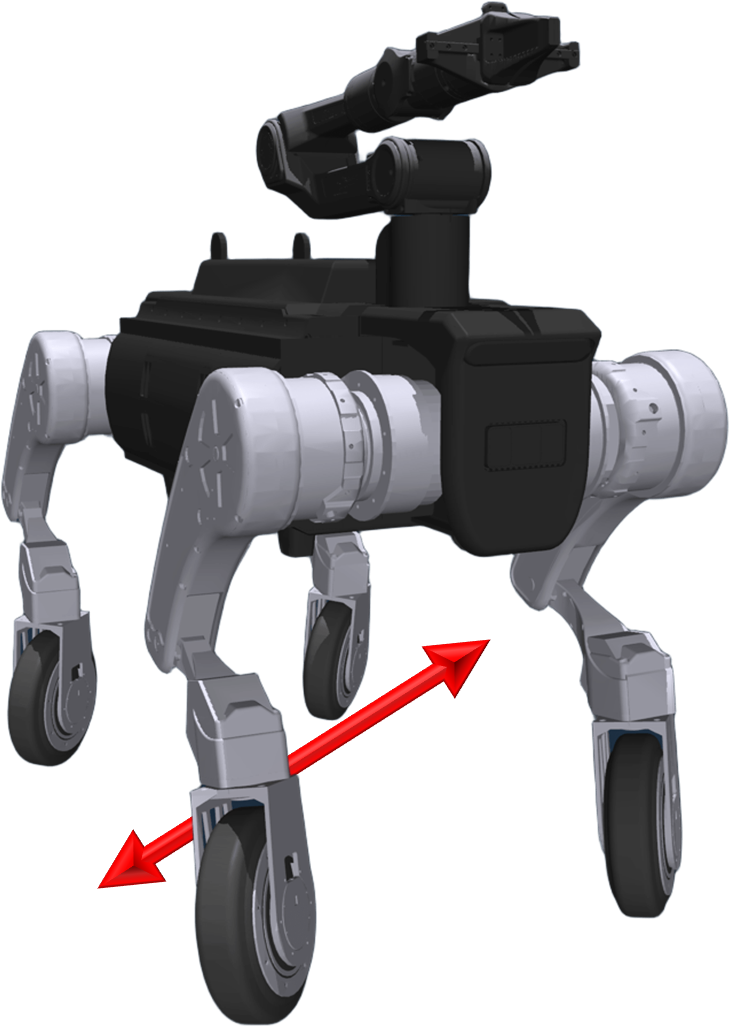}
        \caption{Crab/diagonal motion}
        \label{fig_crab}
    \end{subfigure}
    \begin{subfigure}[b]{0.24\textwidth}
        \includegraphics[width=0.9\textwidth]{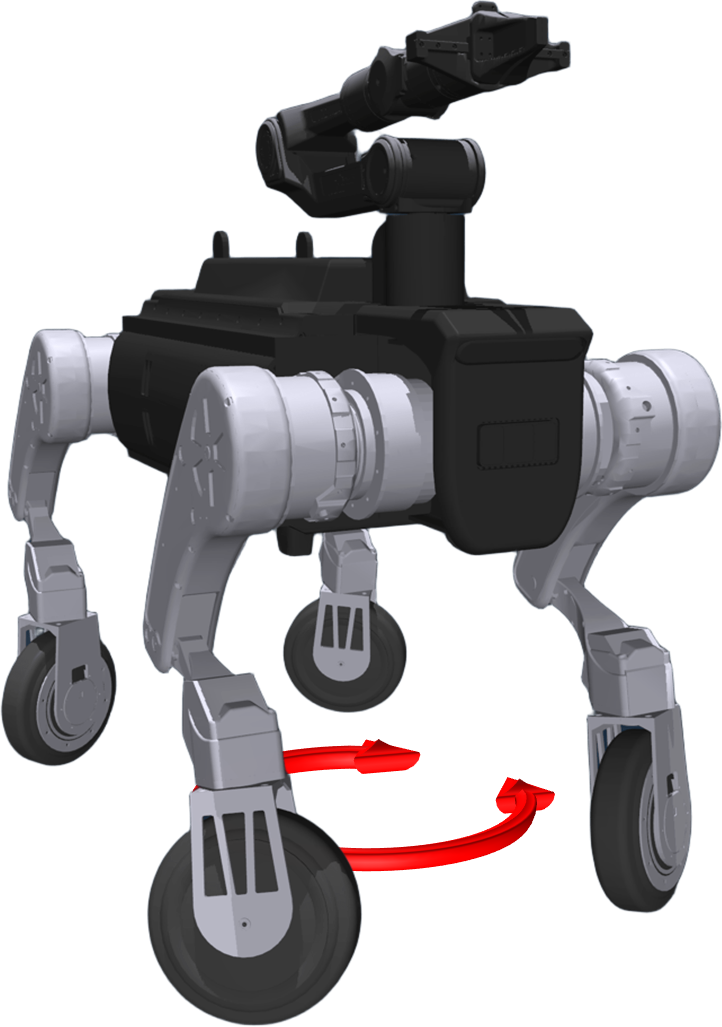}
        \caption{In-place rotation}
        \label{fig_inplace}
    \end{subfigure}
    \begin{subfigure}[b]{0.24\textwidth}
        \includegraphics[width=0.9\textwidth]{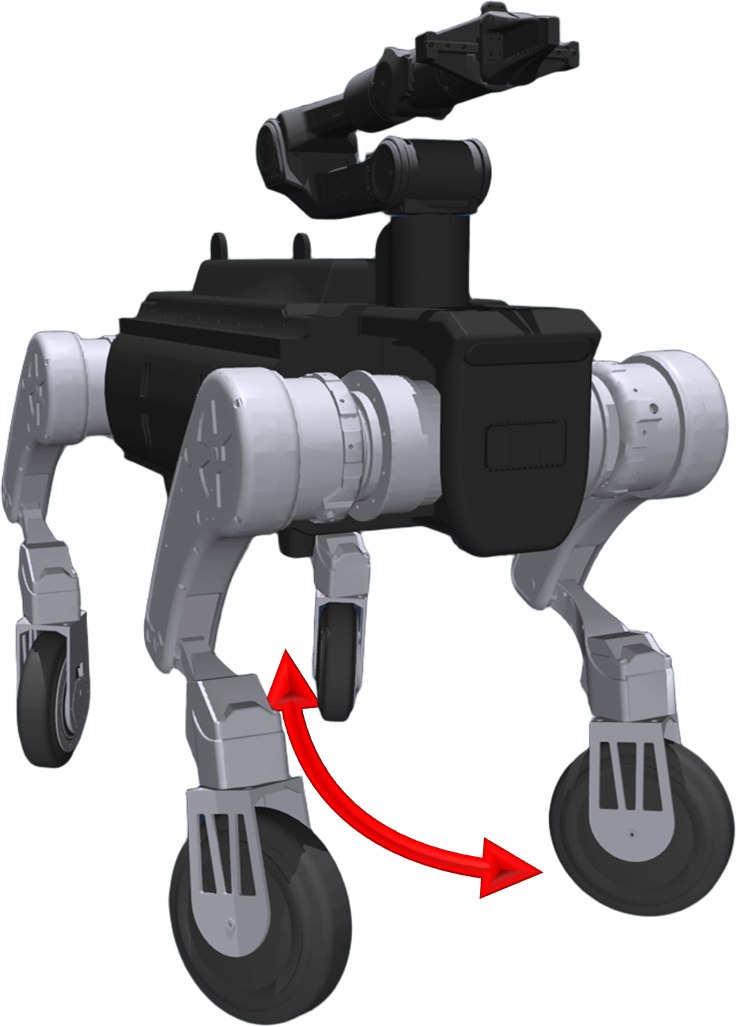}
        \caption{Pivot turning}
        \label{fig_pivot}
    \end{subfigure}
    \caption{Illustration of four basic motion modes for omnidirectional wheel-legged robots}
    \label{fig_steer_wheel}
\end{figure*}

In the proposed omnidirectional quadrupedal wheeled robot, wheel-based locomotion dominates the robot's operational scenarios, particularly in structured, flat environments where mobility efficiency is critical. It is thus necessary to establish a unified kinematic model for wheeled motion to enable efficient task scheduling and seamless control transitions. Since the platform operates mainly at low-to-medium speeds ($ < 2$ m/s), a kinematic model suffices for the required control accuracy. For robots equipped with four-wheel independent steering and four-wheel independent driving (4WIS-4WID), locomotion is typically decomposed into four distinct modes, as illustrated in Fig.~\ref{fig_steer_wheel}: (a) the classical Ackermann steering configuration \cite{chang2024optimal}; (b) crab or diagonal motion, where all wheels steer in the same direction to enable lateral translation \cite{cho2025introduction}; (c) in-place rotation, where wheels rotate around a central axis to achieve zero turning radius \cite{lu2023research}; and (d) pivot turning, where front and rear wheels steer in opposite directions to minimize the turning radius \cite{sun2024disturbance}.
\par
Although these mode-based motion strategies can be effectively deployed in specific tasks, they rely on predefined switching mechanisms and are less adaptive to dynamic or unstructured environments. Most existing works focus on high-level planners to select appropriate motion modes in response to environmental variations\cite{cho2025introduction, chang2024optimal, haider2025advanced}. However, such approaches still exhibit limitations inherent to traditional vehicle control paradigms and fail to fully exploit the motion flexibility of the 4WIS-4WID system.
\par
To address this issue and enable more continuous and unified control strategies, we propose a geometry-aware kinematic modeling approach that captures the full mobility of the omnidirectional platform. For the omnidirectional wheel-legged robot, under the planar, low-speed operating assumption (negligible heave and roll/pitch), the omnidirectional wheel-legged platform is naturally modeled on the configuration manifold \cite{bloch2004nonholonomic},
$$\mathcal{C}_{\text{sw}}= SE(2) \times (\mathbb{S}^1)^4_{\text{steer}} \times (\mathbb{R}^1)^4_{\text{wheel}} \subset \mathbb{R}^{11}$$
where the group factor $SE(2)$ describes the body pose $g=(x_{\text{b}},y_{\text{b}},\theta_{\text{b}})$ and the shape space $M:= (\mathbb{S}^1)^4_{\text{steer}}\times\mathbb{R}^4_{\text{wheel}}$ collects steering angles $q_{\text{steer}}$ and wheel spins $q_{\text{wheel}}$. The left action of $SE(2)$ on $\mathcal{C}_{\text{sw}}$ (here, \{sw\} denotes “steer-wheel”) leaves both the kinetic energy and the rolling constraints invariant, enabling a reduction into “group” and “shape” variables. 
\par
Let $\xi = g^{-1}\dot{g} \in\mathfrak{se}(2)$ be the body twist (left-trivialized velocity) and $u=\dot r\in T_r M$ be the shape velocity, where $r=(q_{\mathrm{steer}},q_{\mathrm{wheel}})$. The rolling-without-slipping and no-lateral-skid constraints define a linear relation between $\xi$ and $u$ that can be written as a principal kinematic connection,
$$\xi = -\mathcal{A}(r)u$$
where $\mathcal{A}(r):T_rM\to\mathfrak{se}(2)$ is the local connection (also called the principal kinematic connection), which maps shape velocities $\dot{r}$ to the body twist $\xi$ via the kinematic reconstruction.
\par
For the 4WIS-4WID system,  each wheel can be independently steered and driven, the connection $\mathcal{A}(r)$ generically has full row rank three (see Appendix), so any planar body velocity $(v_{\text{b,x}},v_\text{b,y},\omega_b)$ can be synthesized by suitable $u$ under the nonholonomic constraints (instantaneous omnidirectionality).
\par
We adopt the reduced state of manifold $\mathcal{C}_{\text{sw}}$ as the system state in the proposed unified kinematic model,
$$X_{\mathrm{SW}}=[x_{\text{b}},y_{\text{b}},\theta_{\text{b}},v_{\text{b,x}},v_{\text{b,y}},\omega_{\text{b}} ]^{\mathrm{T}},$$
where $(v_{\text{b,x}}, v_{\text{b,y}})$ represent the planar linear velocities, and $\omega_{\text{b}}$ denotes the angular velocity about the vertical axis. \par
To ensure smooth, continuous motion (avoiding velocity jumps), we use accelerations as high-level inputs,
$$U_{\mathrm{SW}} = [a_\text{b,x}, a_\text{b,y}, \alpha_{\text{b}}],$$
with $(a_{\text{b,x}}, a_{\text{b,y}})$ the planar accelerations and $\alpha_{\text{b}}$ the angular acceleration.
\par
In low-to-medium speed operation, we adopt a two-layer actuation: steering actuators are position-controlled while wheel motors are velocity-controlled. The actuator-space input is thus defined as,
$$U_{4\mathrm{SW}} = [{q}_{\text{steer}}, \dot{q}_{\text{wheel}}] \in (\mathbb{S}^1)^4 \times \mathbb{R}^4,$$
and is realized from the body-space command via the nonholonomic mapping below.
\par
Taking the geometric center of the robot body as the reference frame, let the position of the $i$-th wheel in the body coordinate frame be $\mathbf{r}_i = [x_i, y_i]^{\mathrm{T}}$, the contact point velocity in the body frame is,
\begin{equation}
    \mathbf{v}_i=\left[ \begin{array}{c}	v_{\mathrm{b},\mathrm{x}}\\	v_{\mathrm{b},\mathrm{y}}\\\end{array} \right] +\omega _bJ\mathbf{r}_i,\,\, J=\left[ \begin{matrix}	0&		-1\\	1&		0\\\end{matrix} \right] 
    \label{eq_wheel_velocity}
\end{equation}
Then the desired steering is the direction of $\mathbf{v}_i$,
\begin{equation}
    \delta _{i}^{\mathrm{opt}}=\mathrm{atan} 2\left( v_{\mathrm{b},\mathrm{y}}+\omega _{\mathrm{b}}x_i, \,\, v_{\mathrm{b},\mathrm{x}}-\omega _{\mathrm{b}}y_i \right)
    \label{eq_wheel_steer_angle}
\end{equation}
and pure rolling gives the wheel spin
$$\dot{\phi}_i=\frac{\left\| \mathbf{v}_i \right\|}{\rho} $$
where, $\rho$ is the wheel radius.
To reduce unnecessary actuator movements and improve control efficiency, the change in steering angle should be minimized. Moreover, given the front-rear symmetry of the steering mechanism, each wheel can be driven in reverse to avoid large-angle rotations. Thus, the optimized steering angle $\delta_i^{\mathrm{opt}}$ is defined as,
\begin{equation}
    \delta_i^\mathrm{opt} = \begin{cases} \delta_i^\mathrm{des} & |\delta_i^\mathrm{des} - \delta_i^\mathrm{cur}| \le \frac{\pi}{2} \\ \delta_i^\mathrm{des} + \pi \ (\bmod\ 2\pi),\quad v_i := -v_i & \text{otherwise} \end{cases}
    \label{eq_wheel_steer_angle_opt}
\end{equation}
Therefore, the complete kinematic mapping from system state to wheel-level control commands is constructed for omnidirectional wheeled robots. Given the system state $X_{\mathrm{SW}}$ and the wheel positions $\{\mathbf{r}_i\}_{i=1}^4$, the kinematic mapping function $\mathcal{F}_{4\mathrm{SW}}$ defines how to compute the control outputs $(v_i, \delta_i^\mathrm{opt})$ for each wheel, where the output space $\mathcal{W}$ is given by,
\begin{equation}
\begin{aligned}
    &\mathcal{F}_{4\mathrm{SW}} : \left( X_{\mathrm{SW}}, \{\mathbf{r}_i\}_{i=1}^4 \right) \mapsto \mathcal{W}, \\
    \mathcal{W} &= \left\{(v_i, \delta_i^\mathrm{opt}) \in \mathbb{R} \times \mathbb{S}^1 \mid i = 1, \dots, 4 \right\}
\end{aligned}
\label{eq_wheel_kinematics}
\end{equation}

\subsection*{Formulation of Motion Control}
Based on the previously established multibody dynamics model and the omnidirectional wheeled kinematic model, the whole-body motion control of the wheel-legged robot is formulated as a constrained optimization problem. Under the constraints imposed by system dynamics and contact conditions, the objective is to optimize a cost function to enhances motion performance and task execution accuracy. For wheel-legged robots with both legged and wheeled locomotion capabilities, multiple feasible strategies may exist to reach a target pose—either through wheeled locomotion or legged stepping. This redundancy can result in non-unique or inconsistent optimization outcomes. To ensure a physically meaningful solution and improve control robustness, we explicitly differentiate and constrain the two primary locomotion modes within the control framework,
\begin{itemize}
    \item Wheeled Locomotion Mode: All wheels remain in continuous contact with the ground. The steering joints and wheel hub motors are actively controlled to track the desired motion. The robot body remains stable without leg lifting or swinging, this mode is suitable for fast and efficient locomotion on flat or mildly uneven terrain.
    \item Legged Locomotion Mode: All steering joints and wheel motors are locked, and each leg acts as a rigid kinematic chain to perform swing or stance motions. In this mode, wheels provide only passive contact or support. This strategy is used to traverse obstacles or complex unstructured terrain.
\end{itemize}
\par
Under a finite-horizon discrete-time optimal control framework, the whole-body motion control problem for the wheel-legged robot with arm can be formalized as the following optimization problem,
\begin{equation}
\begin{aligned}
\left\{ 
    \begin{array}{c}
        \mathrm{s}_{0}^{*},\cdots ,\mathrm{s}_{N}^{*}\\
        \mathrm{u}_{0}^{*},\cdots ,\mathrm{u}_{N-1}^{*}
    \end{array}
\right\} 
&= \mathrm{arg} \,\, \underset{\mathbf{S},\mathbf{U}}{\min}\,\,\sum_{k=0}^{N-1}{\ell}(\mathrm{s}_k,\mathrm{u}_k)\;+\;\ell _f(\mathrm{s}_N) \\
\text{s.t.} \quad \mathrm{s}_{k+1} &= f\left( \mathrm{s}_k,\mathrm{u}_k \right)
\end{aligned}
\label{eq_motion_control_optimization}
\end{equation}
where $\mathrm{s}_k = (\mathrm{q}_k, \dot{\mathrm{q}}_k) \in \mathbb{R}^{n_s}$ represents the system state vector, $\mathrm{u}_k \in \mathbb{R}^{n_u}$ denotes the control input, and $f(\cdot)$ is the dynamics function governed by the contact-constrained multibody system as defined in Equation \eqref{eq_robot_dynamics_contact}. The function $\ell(\mathrm{s}_k,\mathrm{u}_k)$ denotes the stage cost, and $\ell_f(\mathrm{s}_N)$ represents the terminal state cost.
\par
To accommodate the motion characteristics and task requirements of omnidirectional wheel-legged robots, the stage cost $\ell(\cdot)$ and terminal cost $\ell_f(\cdot)$ in the optimization formulation comprehensively incorporate multiple objectives,
\begin{equation}
    \ell (\mathrm{s}_k)=\ell _{k,\mathrm{leg}}^{w_{\mathrm{leg}}}+\ell _{k,\mathrm{CoM}}^{w_{\mathrm{CoM}}}+\ell _{k,f_{\mathrm{c}}}^{w_{f_{\mathrm{c}}}}+\ell _{k,\mathrm{reg}}^{w_{\mathrm{reg}}}+\ell _{k,\mathrm{pos}}^{w_{\mathrm{pos}}}+\ell _{k,\sigma}^{w_{\sigma}}+\ell _{k,\mathrm{arm}}^{w_\mathrm{arm}}
    \label{eq_cost_function}
\end{equation}
here, the superscript $w_*$ indicates the weight assigned to each component. Specifically, $\ell _{*,\mathrm{leg}}$ penalizes leg swing trajectory tracking errors; $\ell _{*,\mathrm{CoM}}$ penalizes deviations from the desired center-of-mass trajectory; $\ell _{*,f_{\mathrm{c}}}$ represents penalties related to external forces and contact; $\ell _{*,\mathrm{reg}}$ encourages control regularity and smoothness; $\ell _{*,\mathrm{pos}}$ penalizes deviations from the desired body orientation; $\ell _{*,\sigma}$ penalizes steering joint posture deviations; and $\ell _{*,\mathrm{arm}}$ constrains the end-effector posture of the manipulator arm.
\par
According to the Bellman optimality principle, we define the value function $V_k(\mathrm{s})$ as the minimum accumulated cost from state $\mathrm{s}$ at time step $k$ to the final step $N$ under the optimal control policy,
\begin{equation}
\begin{aligned}
    V_k(\mathrm{s}) \;=\;&\min_{\{\mathrm{u}_i\}_{i=k}^{N-1}}\;\sum_{i=k}^{N-1}\ell(\mathrm{s}_i,\mathrm{u}_i)+\ell_f(\mathrm{s}_N) \\
    	&\text{s.t.}\quad \mathrm{s}_{i+1} = f\!\left( \mathrm{s}_i,\mathrm{u}_i \right)
\end{aligned}
\end{equation}
The value function satisfies the following recursive relation, known as the Bellman equation,
\begin{equation}
    V_k(\mathrm{s}) = \min_{\mathrm{u}_k} \left\{ \ell(\mathrm{s}_k,\mathrm{u}_k) + V_{k+1}(f(\mathrm{s}_k,\mathrm{u}_k)) \right\}
    \label{eq_value_function}
\end{equation}
This transforms the original high-dimensional optimal control problem \eqref{eq_motion_control_optimization} into a sequence of nested dynamic programming subproblems.
\par
However, due to the general nonlinearity of the dynamics function $f(\mathrm{s},\mathrm{u})$ and the value function $V_k(\mathrm{s})$, the Bellman equation is difficult to solve directly in practice. A common approach is to approximate the value function locally via second-order Taylor expansion around the current trajectory $(\mathrm{s}_k, \mathrm{u}_k)$ , forming a local linear-quadratic (LQ) subproblem. Let the deviation variables be defined as $\delta \mathrm{s} = \mathrm{s} - \mathrm{s}_k$ and $\delta \mathrm{u} = \mathrm{u} - \mathrm{u}_k$, The increment of the value function can then be expressed as,
\begin{equation}
    \begin{aligned}
    Q_k(\delta \mathrm{s},\delta \mathrm{u})\;=\; \underbrace{\ell(\mathrm{s}_k,\mathrm{u}_k)+V_{k+1}(\mathrm{s}_{k+1})}_{\text{const}} + \begin{bmatrix}Q_\mathrm{s}\\Q_\mathrm{u}\end{bmatrix}^\text{T} \begin{bmatrix}\delta \mathrm{s}\\\delta \mathrm{u}\end{bmatrix} \\
     + \tfrac12 \begin{bmatrix}\delta \mathrm{s}\\\delta \mathrm{u}\end{bmatrix}^\text{T} \begin{bmatrix}Q_{\mathrm{s}\mathrm{s}}&Q_{\mathrm{s}\mathrm{u}}\\Q_{\mathrm{u}\mathrm{s}}&Q_{\mathrm{u}\mathrm{u}}\end{bmatrix} \begin{bmatrix}\delta\mathrm{s}\\\delta \mathrm{u}\end{bmatrix}
    \end{aligned}
\end{equation}
where, $Q_k(\delta \mathrm{s},\delta \mathrm{u})$ is the second-order local approximation of the value increment at time step $k$, and the coefficients are defined as
$$
\begin{aligned} Q_\mathrm{s} &= \ell_\mathrm{s} + f_\mathrm{s}^\text{T} V'_{\mathrm{s}},& Q_\mathrm{u} &= \ell_\mathrm{u} + f_\mathrm{u}^\text{T} V'_{\mathrm{s}},\\ Q_{\mathrm{s}\mathrm{s}} &= \ell_{\mathrm{s}\mathrm{s}} + f_\mathrm{s}^\text{T} V'_{\mathrm{s}\mathrm{s}}\,f_\mathrm{s},& Q_{\mathrm{s}\mathrm{u}} &= \ell_{\mathrm{s}\mathrm{u}} + f_\mathrm{s}^\text{T} V'_{\mathrm{s}\mathrm{s}}\,f_\mathrm{u},\\ Q_{\mathrm{u}\mathrm{u}} &= \ell_{\mathrm{u}\mathrm{u}} + f_\mathrm{u}^\text{T} V'_{\mathrm{s}\mathrm{s}}\,f_\mathrm{u}& \end{aligned}
$$
here, $f_\mathrm{s}$ and $f_\mathrm{u}$ are the Jacobians of the dynamics function $f(\cdot)$ with respect to the state and control, respectively. $V'_{\mathrm{s}}$ and $V'_{\mathrm{s}\mathrm{s}}$ denote the gradient and Hessian of the next-step value function $V_{k+1}(\mathrm{s}_{k+1})$.
\par
Based on the local second-order approximation of the incremental value function $Q_k(\delta \mathrm{s},\delta \mathrm{u})$, the optimal control increment $\delta \mathrm{u}_k^*$, that minimizes the local cost at each time step $k$can be obtained analytically as,
\begin{equation}
    \delta \mathrm{u}_k^* = \arg \min_{\delta \mathrm{u}_k} Q_k(\delta \mathrm{s}_k, \delta \mathrm{u}_k) = k_k + K_k \delta \mathrm{s}_k
    \label{eq_optimal_increment}
\end{equation}
here, $k_k = -Q_\mathrm{u} Q_{\mathrm{u}\mathrm{u}}^{-1}$ is the feedforward term, and $K_k = -Q_{\mathrm{s}\mathrm{u}} Q_{\mathrm{u}\mathrm{u}}^{-1}$ is the feedback gain matrix. This control law provides a locally optimal update direction in the vicinity of the current state deviation $\delta \mathrm{s}_k$, effectively guiding the system state to converge toward the desired trajectory. \par
\begin{figure*}[t]
    \centering
    \includegraphics[width=0.98\textwidth]{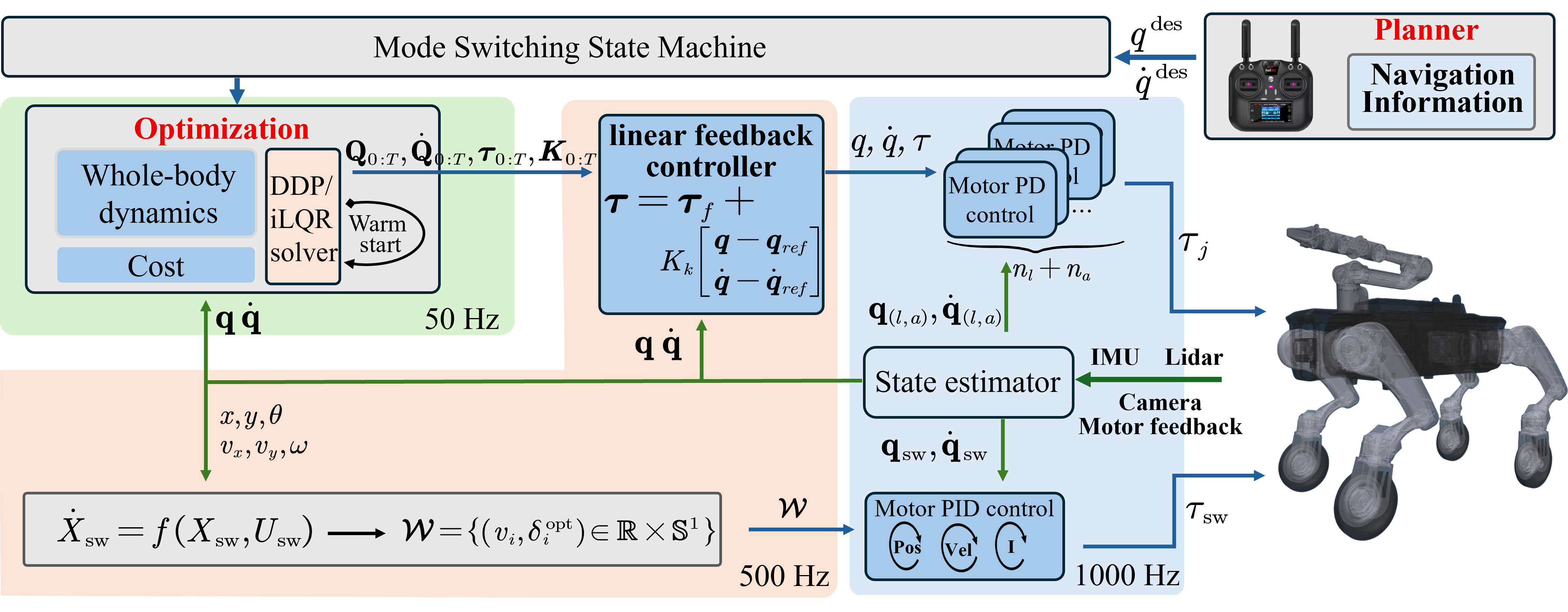}
    \caption{The motion control framework of the omnidirectional wheeled quadruped robot}
    \label{fig_framework}
\end{figure*}
The value function at the current time step is also updated using first- and second-order approximations as follows,
$$
V_\mathrm{s} = Q_\mathrm{s} - K_k^\text{T} Q_{\mathrm{u}\mathrm{u}} k_k, \quad V_{\mathrm{s}\mathrm{s}} = Q_{\mathrm{s}\mathrm{s}} - K_k^\text{T} Q_{\mathrm{u}\mathrm{u}} K_k
$$
Starting from the terminal value function initialized as $V_N = \ell_f(\mathrm{s}_N)$, a backward pass is performed from time step $k = N - 1$ to compute the full sequence of locally optimal feedback policies $\boldsymbol{k}_{0:T},\boldsymbol{K}_{0:T}$. A subsequent forward pass simulates the system dynamics by rolling out the new control policy to generate an updated state-control trajectory $(\mathrm{s}_k^{\text{new}}, \mathrm{u}_k^{\text{new}})$, with optional line search applied to ensure convergence and step size regularization. This process is repeated iteratively until convergence criteria are met.
\par
During DDP-based optimization, the feedback gain matrix ${K}_k$ at each time step yields a full-body feedback control law. At the low-level control layer, a high-frequency Linear Feedback Controller (LFC) can utilize the state deviation $\delta \mathbf{s}_k = \mathbf{s}_k - \mathbf{s}_k^\mathrm{des}$ and the corresponding gain ${K}_k$ to achieve real-time stabilization and enhance dynamic responsiveness. Furthermore, the use of Feasibility-Driven Differential Dynamic Programming (FDDP) \cite{crocoddyl20icra}, or its robust variant Box-FDDP \cite{mastalli2022feasibility}, improves convergence when initialized with poor guesses, thereby enhancing the framework's robustness to disturbances and adaptability to varying operating conditions.
\par
As illustrated in Fig.~\ref{fig_framework}, we propose a unified contact motion control framework for omnidirectional wheel-legged robots. The framework begins with a Mode Switching State Machine, which selects the current locomotion mode based on task commands, and generates the corresponding reference trajectory from high-level navigation planning or teleoperation. The reference trajectory is passed to the Optimal Control Module, which formulates the motion control problem as a constrained finite-horizon optimal control problem based on the contact dynamics model described in Equation~\eqref{eq_robot_dynamics_contact}, and the associated cost and constraint terms given in Equation~\eqref{eq_cost_function}. The optimization problem follows the formulation in Equation~\eqref{eq_motion_control_optimization}, and is solved efficiently using DDP-based solvers.

\par
The outputs of the optimization module include the optimal state trajectory $\mathbf{S} = [\mathbf{q}^\text{T}, \dot{\mathbf{q}}^\text{T}]^\text{T}$, the optimal control torques $\boldsymbol{\tau}$, and the time-varying full-body feedback gain matrices ${K}_k$. These outputs are passed to the Linear Feedback Controller (LFC), which operates at a frequency of 500 Hz, using the discrete time interval $\Delta t$ specified during the DDP optimization process. At each step, it computes a feedback torque correction based on the gain matrix ${K}_k$ and the deviation between the current and desired states. The resulting torque $\boldsymbol{\tau}$, along with the actual joint states $\mathbf{q}, \dot{\mathbf{q}}$, is forwarded to the underlying motor-level PD controller to improve responsiveness and prevent instability such as motor overspeed.
\par
In the wheeled control, based on the state $X_{\mathrm{SW}}$ and input $U_{\mathrm{SW}}$, as defined previously, the system first computes the desired base velocity command $(v_{\text{b},x}^*, v_{\text{b},y}^*, \omega_{\text{b}}^*)$. It then analytically calculates the steering angles and wheel speeds using Equations~\eqref{eq_wheel_velocity}, \eqref{eq_wheel_steer_angle}, and \eqref{eq_wheel_steer_angle_opt}, and sends these commands to the corresponding motor controllers for execution in position and velocity control modes, respectively.
\par
During whole-body coordination, the robot's wheel and leg structure adapts to the selected mode: In wheeled mode, the leg joints remain fixed, ensuring continuous ground contact and stable traction; in legged mode, the wheels are locked, steering and drive joints are frozen, and the legs perform gait or jumping motions. This prevents motion errors caused by unintended wheel rotation.
\par
Regarding optimization efficiency, the DDP algorithm solves a sequence of local quadratic subproblems for fast convergence but remains sensitive to initial trajectory guesses. For our wheel-legged system with 41 state variables and 160 time steps, $cold \,\, start$ optimization (zero initial guess) on the Nvidia Jetson Orin NX platform requires approximately 130-170 ms to converge to a precision of $10^{-4}$. By adopting a $warm \,\, start$ strategy—using the previous optimization result as the initial guess—and reducing the number of nodes to 20, the computation time during execution is reduced to 11-13 ms, and can reach as low as 3-4 ms in stable walking phases, meeting the requirements of high-frequency real-time control.
\section*{Results}
\begin{figure*}[htbp]
    \centering
    \includegraphics[width=1\textwidth]{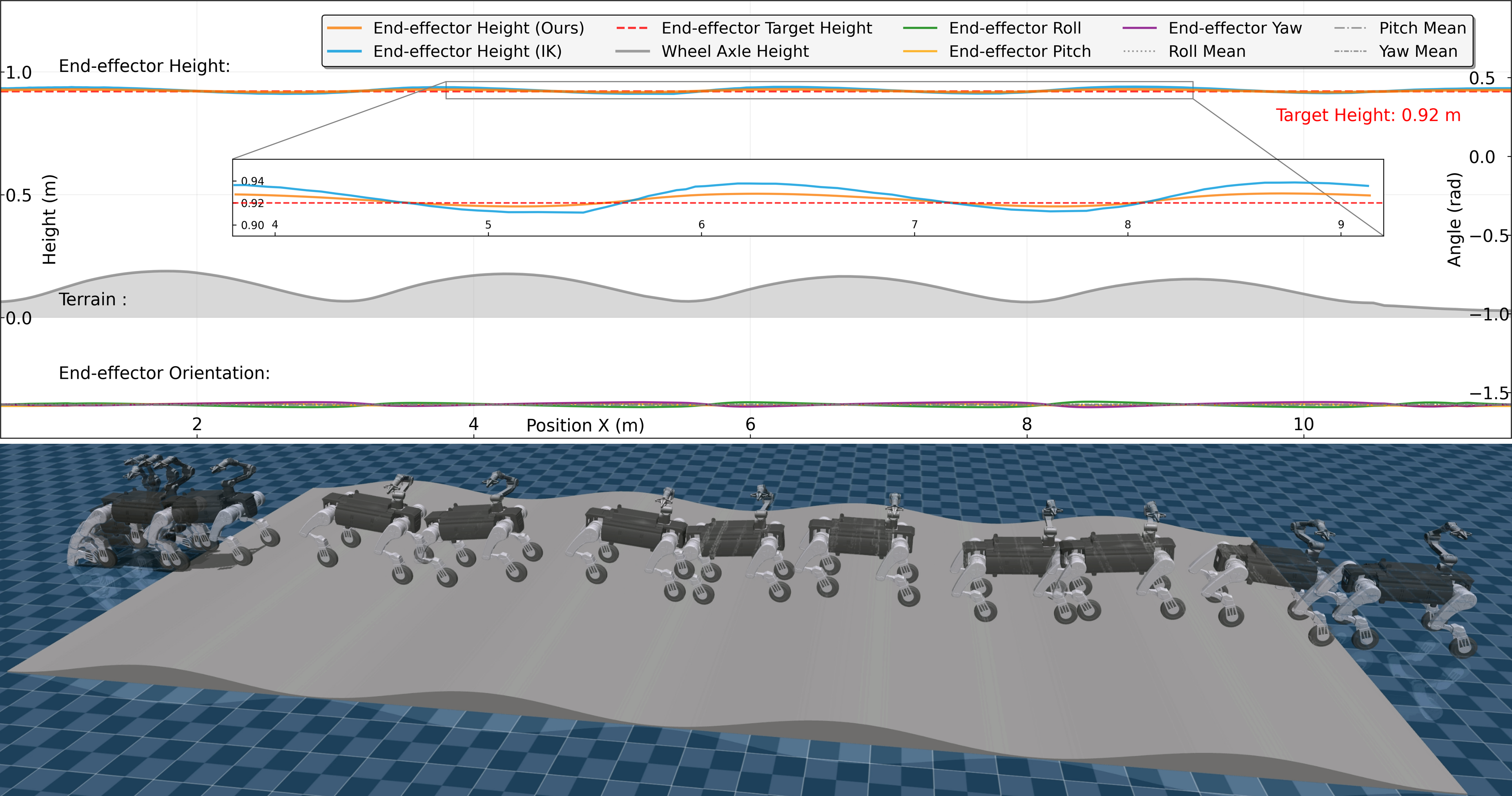}
    \caption{Simulation scenario: the omnidirectional wheel-legged robot equipped with a manipulator traverses a wave-shaped terrain. The ground length is 11.5 m, peak height 0.2 m, end-effector target height is 0.92 m, and orientation target is (-1.57, 0, -1.57) rad in RPY format. The pitch angle is shifted by -1.57 rad in the figure for clarity. The end-effector orientation is plotted on the right axis; other variables use the left axis.}
    \label{fig_sim_terrain}
\end{figure*}
In this section, we conduct comprehensive evaluations of the proposed whole-body motion control framework for omnidirectional wheel-legged robots through three simulation scenarios and two sets of hardware experiments. The goal is to assess the system's stability and control performance under diverse tasks. As shown in Fig.~\ref{fig_robot_model}, the experimental platform comprises 8 leg joints, 4 steering joints, 4 wheel-driving joints, and a 7-Dof manipulator equipped with a gripper. The three simulation environments are conducted using MuJoCo, and illustrated in Figs.~\ref{fig_sim_terrain},~\ref{fig_sim_fixed}, and~\ref{fig_sim_trans}, while the physical robot experiments are shown in Fig.~\ref{fig_hw_Experiments}. Across all tasks, the control objective is to maintain the manipulator end-effector's height and the orientation as stable as possible.
\subsection*{Simulation}
\begin{figure*}[htbp]
    \centering
    \begin{subfigure}[b]{0.495\textwidth}
        \includegraphics[width=\linewidth, height=2in]{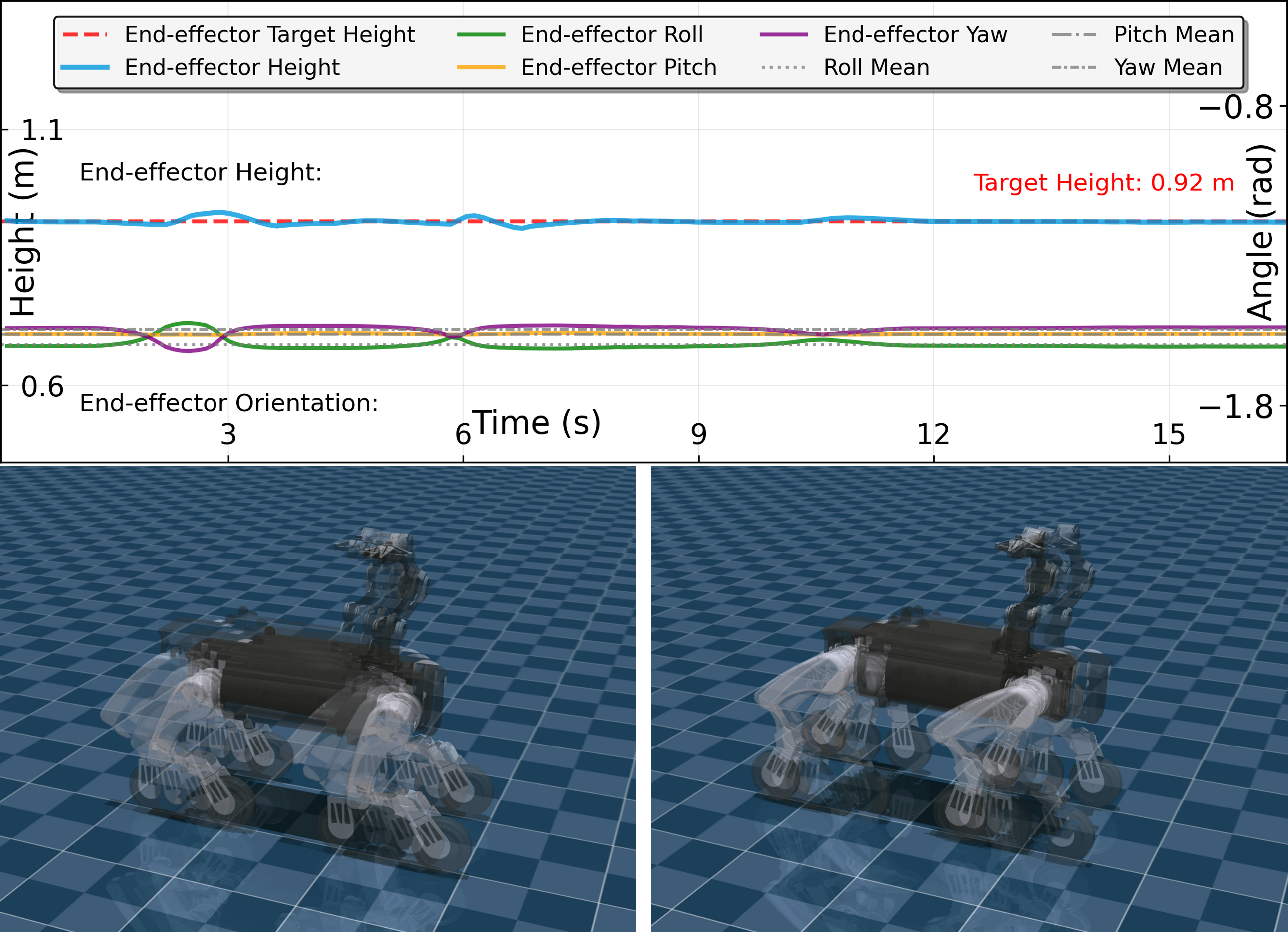}
        \caption{Fixed wheel simulation}
        \label{fig_sim_fixed}
    \end{subfigure}
    \hfill 
    \begin{subfigure}[b]{0.495\textwidth}
        \includegraphics[width=\linewidth, height=2in]{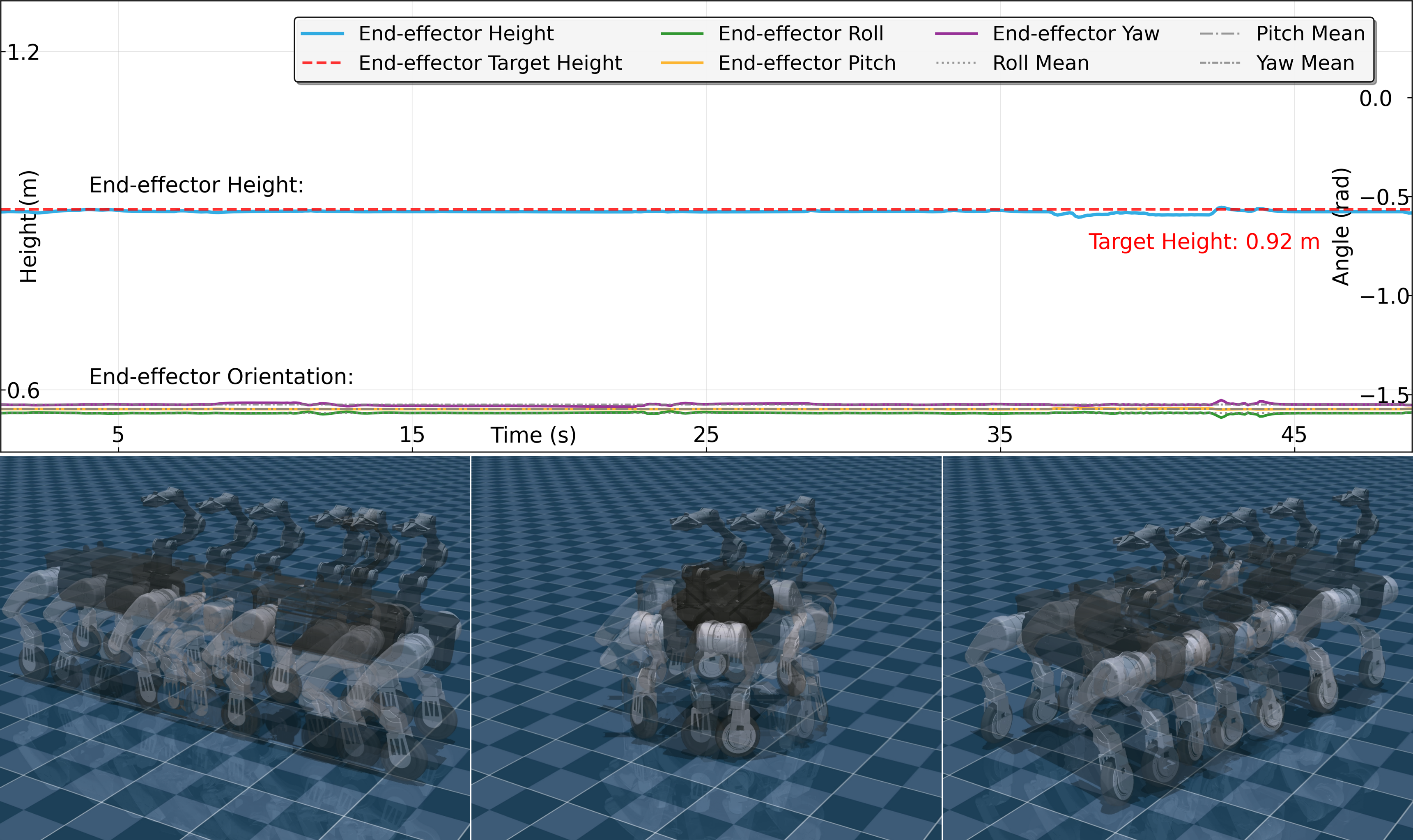}
        \caption{Trans wheel simulation}
        \label{fig_sim_trans}
    \end{subfigure}
    \caption{Simulation results on flat terrain with a manipulator-equipped omnidirectional wheel-legged robot. (a) Base-fixed test under wheel perturbation. (b) Wheeled locomotion tests under various gaits. The end-effector orientation is plotted on the right axis; other variables use the left axis.}
    \label{fig:2}
\end{figure*}

Three representative scenarios were designed in MuJoCo, with the end-effector target height set to 0.92 m. (1) \textit{Uneven terrain traversal}: The robot traverses a wave-shaped terrain while maintaining the manipulator end-effector's stability, as shown in Fig.~\ref{fig_sim_terrain}. (2) \textit{Fixed base test}: On flat ground, the robot's base is fixed while forward and backward wheel speeds are applied. The relative pose between the base and the manipulator is observed (Fig.~\ref{fig_sim_fixed}). (3) \textit{Wheeled locomotion modes}: On flat terrain, different wheeled motion patterns such as crab walking and in-place spinning are simulated to verify full-body motion control and end-effector stability (Fig.~\ref{fig_sim_trans}). 
\par
As shown in Fig.~\ref{fig_sim_terrain}, a ghosted trajectory overlay illustrates the robot's motion as it traverses the undulating terrain. We compare our proposed whole-body control approach with a conventional inverse kinematics (IK)-based arm control method in terms of end-effector height stability. From the magnified view, it is evident that our whole-body control method significantly reduces oscillations in the end-effector during terrain traversal compared to the IK-based approach. The underlying reason is that, in the IK-based method, the base pose used for arm inverse kinematics is derived from the robot's center of mass (CoM) estimate. However, when the robot body travels over uneven terrain, the base under the manipulator experiences pose changes before the CoM. In contrast, the proposed method directly incorporates the full-body dynamics into optimization, inherently modeling joint interactions, enabling better adaptation to terrain variations and maintaining end-effector stability. During the entire process of traversing the undulating terrain, using the inverse kinematic solution method, the mean and standard deviation of end-effector height are: $0.928 \ (\pm 0.041)$ m, with a maximum error from the target height is 0.083 m. In the method based on the overall dynamics control framework we proposed, the mean and standard deviation of end-effector height are: $0.924 \ (\pm 0.008)$ m, with a maximum error from the target height is 0.007 m. For the end-effector attitude, the mean and standard deviation of Roll, Pitch, and Yaw are: $-1.578 \ (\pm 0.009)$ rad, $-1.580 \ ( \pm 0.05)$ rad, $ -1.575 \ (\pm 0.08 )$ rad respectively.
\par
\subsection*{Experiments}
\begin{figure*}[h]
    \centering
    \begin{subfigure}{0.496\textwidth}
        \includegraphics[width=1\textwidth]{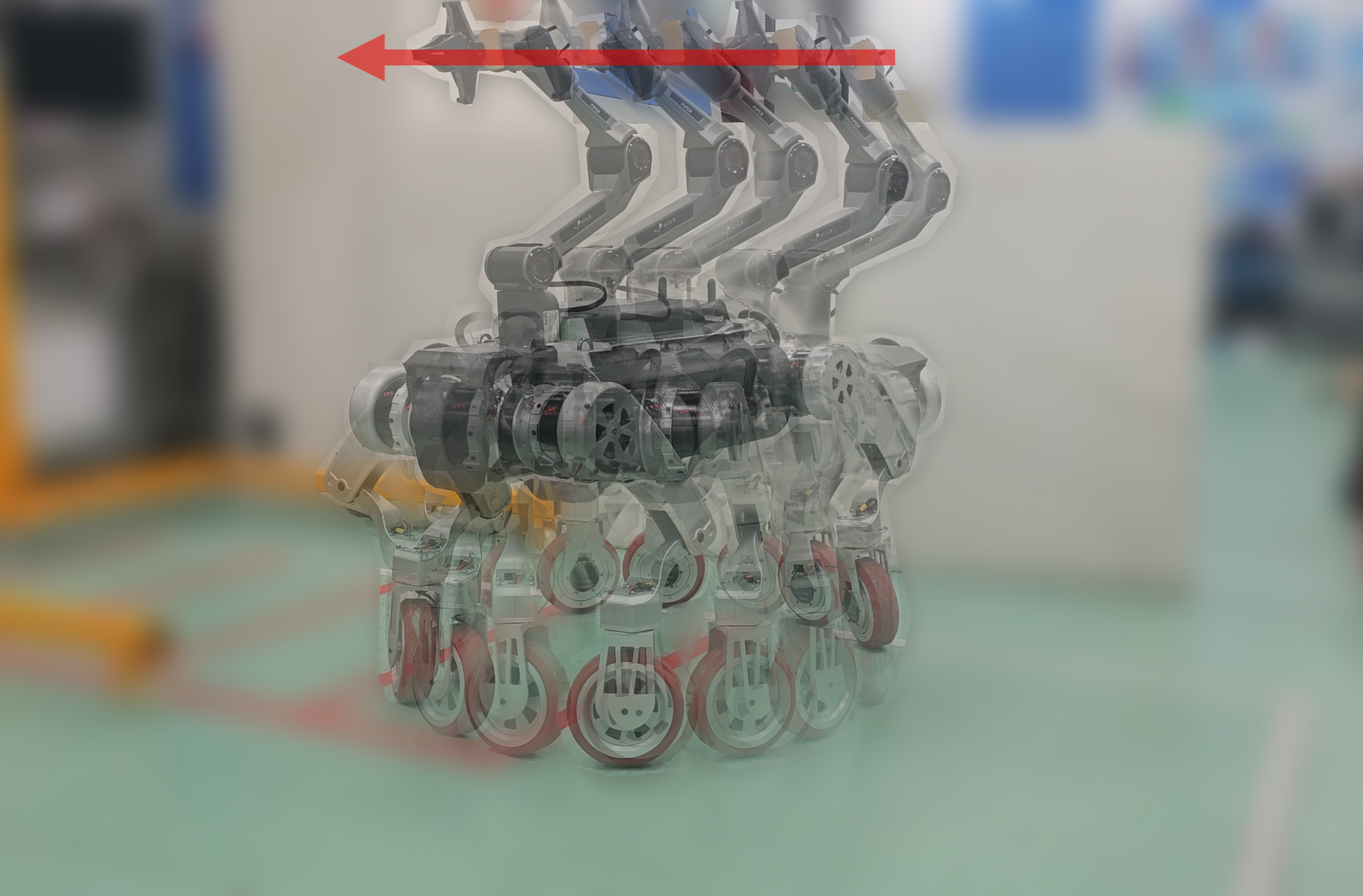}
        \caption{In-place rotation}
        \label{fig_inplace_hw}
    \end{subfigure}
    \begin{subfigure}{0.496\textwidth}
        \includegraphics[width=1\textwidth]{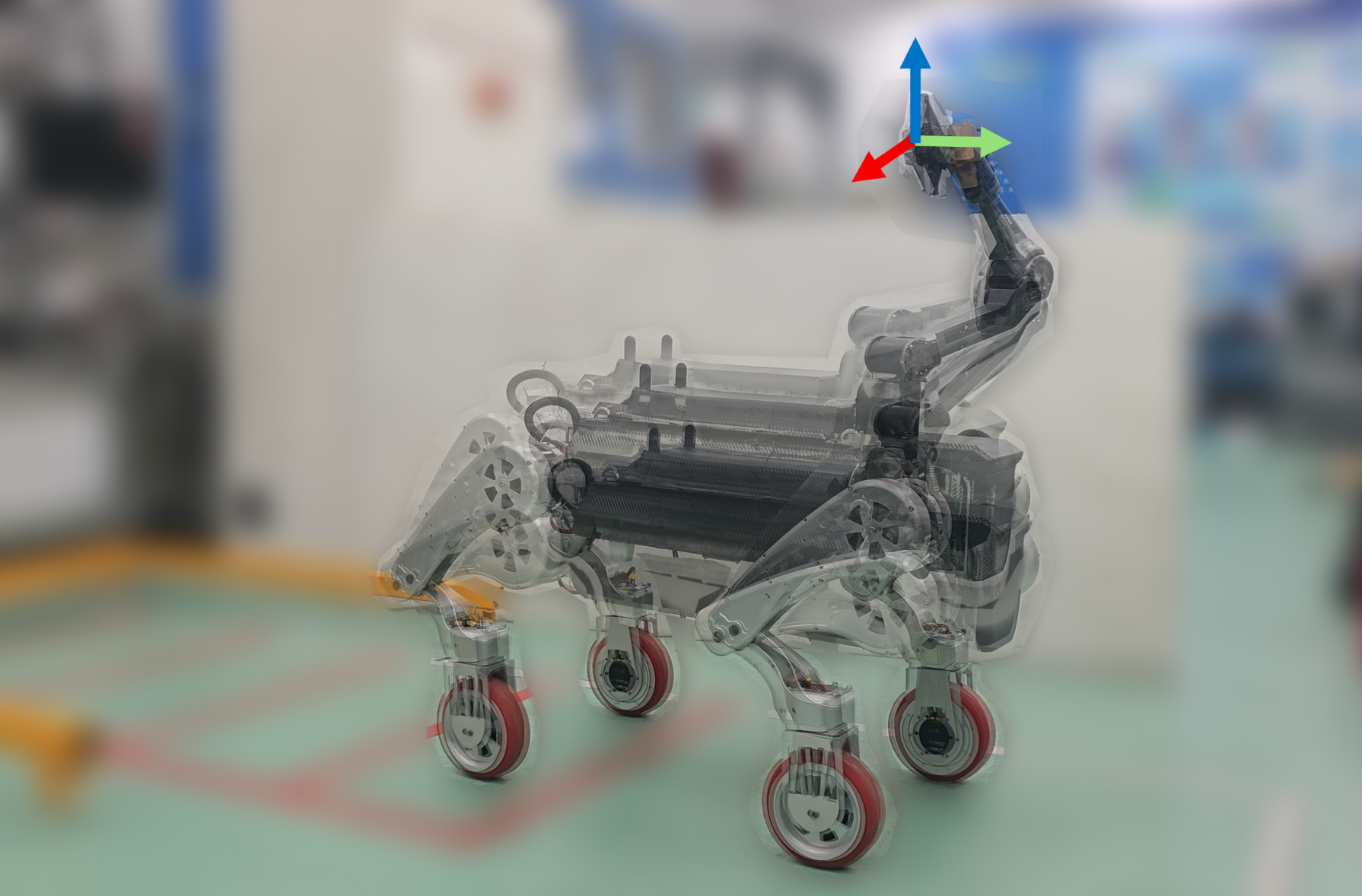}
        \caption{Up-down motion}
        \label{fig_updown_hw}
    \end{subfigure}
    \caption{Hardware experimental results of the omnidirectional wheel-legged robot. Left: end-effector remains stable during in-place rotation. Right: end-effector stabilization during squatting and standing.}
    \label{fig_hw_Experiments}
\end{figure*}
As shown in Fig.~\ref{fig_sim_fixed}, we simulate the omnidirectional wheel-legged robot maintaining a stable upright posture. Starting from a standing state, forward and backward initial velocities are applied to the four wheels to emulate external disturbances acting on the body. The results indicate that, in order to maintain base stability, joint motors exhibit significant angular deviations, with the end-effectors of all four legs shifting forward or backward accordingly. During this motion, the mean and standard deviation of the manipulator end-effector height were measured as $0.919 \ (\pm 0.005)$ m, with a maximum deviation of 0.018 m. The end-effector orientation (Roll, Pitch, Yaw) had means and standard deviations are: $-1.597 \ (\pm 0.016)$ rad, $-1.561 \ (\pm 0.003)$ rad, and $ -1.545 \ (\pm 0.016)$ rad, respectively. In Fig.~\ref{fig_sim_trans}, we evaluate different wheeled locomotion patterns including straight-line driving, in-place rotation, and crab-walking. The end-effector height maintained a mean and standard deviation of $ 0.916 \ (\pm 0.002)$ m, with a maximum deviation of 0.011 m. The corresponding end-effector orientation statistics were, Roll: $ -1.594 \ (\pm 0.033)$ rad, Pitch: $-1.573 \ (\pm 0.009)$ rad, Yaw: $-1.551 \ (\pm 0.057)$ rad.
\begin{figure*}[htbp] 
    \centering
    \includegraphics[width=1\textwidth]{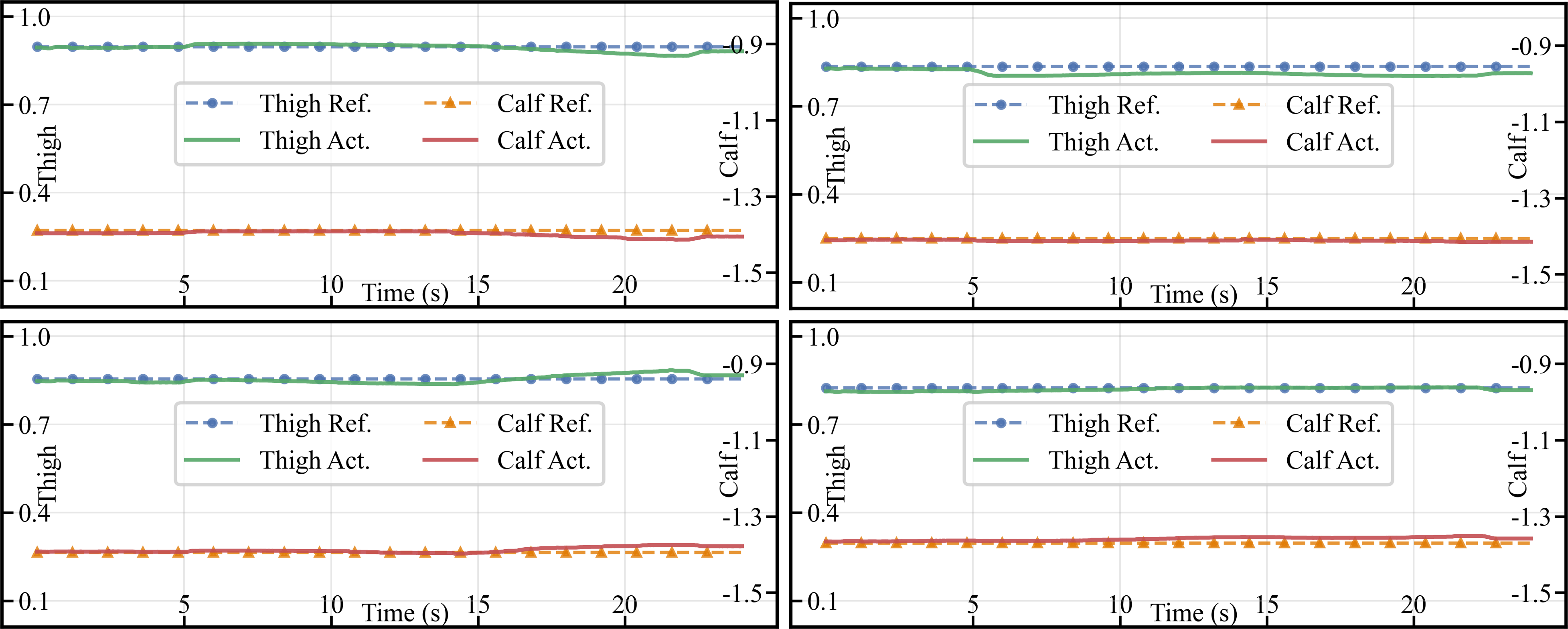}
    \caption{Tracking performance of the joint trajectories during the in-place rotation experiment of the omnidirectional wheel-legged robot. Top-left: front-left leg; top-right: front-right leg; bottom-left: rear-left leg; bottom-right: rear-right leg. Left axis: hip joint; right axis: knee joint.}
    \label{fig_leg_joint}
\end{figure*}
\begin{figure*}[htbp]
    \centering
    \includegraphics[width=1\textwidth]{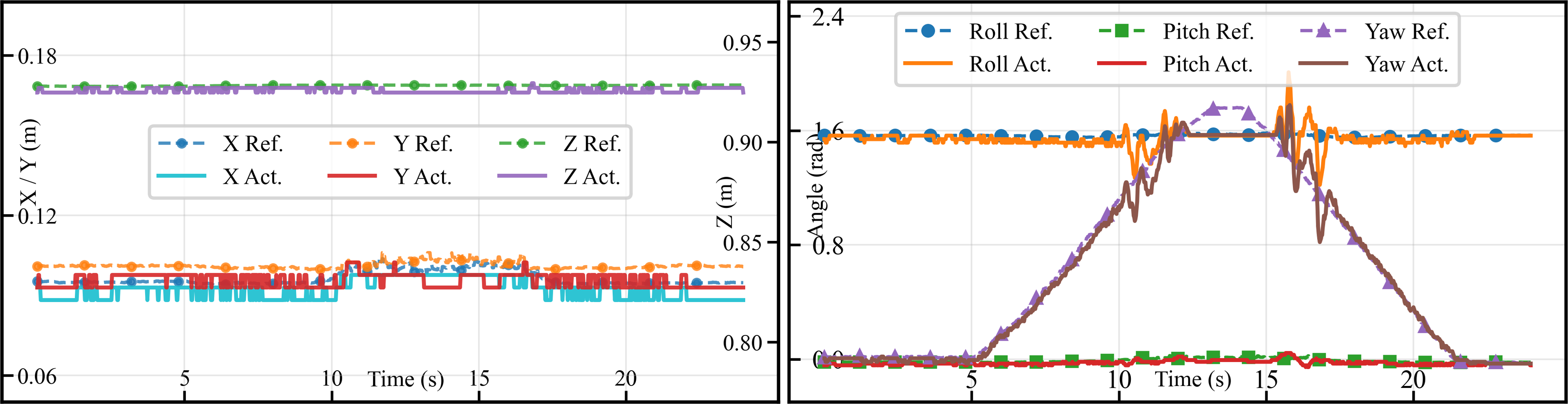}
    \caption{Manipulator end-effector trajectory tracking results during the in-place rotation experiment. Left: comparison of end-effector position (X/Y displacement aligned with left axis, Z-height with right axis). Right: comparison of end-effector orientation in the body-fixed coordinate frame.}
    \label{fig_ee_xyzrpy}
\end{figure*}
To further validate the proposed motion control framework on real hardware, we conducted experiments using our custom-built omnidirectional wheel-legged robotic platform. Two representative scenarios were tested: in-place rotation and squatting-to-standing transition. During both tasks, the manipulator end-effector was required to maintain a preset height and orientation to evaluate the coordination and control precision of the full-body system.
\par
As illustrated in Fig.~\ref{fig_hw_Experiments}, the robot performed an in-place rotation experiment, rotating approximately 1.8 rad about its base's vertical (Z) axis and then returning to the initial orientation. Throughout the motion, the manipulator end-effector successfully maintained the target height of 0.92 m and target orientation of (1.57, 0, 0) rad. The desired trajectories and measured joint angles of the leg joints are shown in Fig.~\ref{fig_leg_joint}, where hip joint data are aligned to the left axis and knee joint data to the right axis. During the spinning motion, the leg joints worked in coordination to stabilize the body pose. The end-effector's position variation during the motion is plotted in the left subfigure of Fig.~\ref{fig_ee_xyzrpy}. The vertical (Z) direction remains nearly constant, verifying that the controller maintained vertical stability. In the horizontal (X/Y) directions, some minor lateral drifts occurred due to compliance in constraints and base movement. Slight sawtooth-like oscillations in the sensor readings were observed, which are attributed to the resolution limit of the hardware sensors, especially during nearly stationary conditions.
\par
The right subfigure of Fig.~\ref{fig_ee_xyzrpy} presents the end-effector's orientation changes over time, expressed in the robot base coordinate frame to emphasize rotational dynamics. A clear rotational and recovery motion in yaw is evident. Notably, two intervals of increased oscillations occurred around 10-12 s and 15-17 s. These oscillations are attributed to a combination of factors: (1) transient disturbances in base pose estimation affecting the end-effector's relative orientation calculation; (2) physical joint limits reached by the manipulator, resulting in abrupt mechanical constraints. Between 12 and 15 s, the manipulator encountered a hard joint limit, leading to significant deviations between actual and desired orientations caused by joint range saturation and diminished controllability near singular configurations.
\section*{Conclusion}
In this paper a full-body motion control framework based on contact-aware dynamics modeling is constructed. The proposed framework incorporates both articulate multi-body dynamics and omnidirectional wheeled kinematics into a unified formulation. By a discrete-time optimal control task over a finite prediction horizon, the method casting the problem recursively as achieves coordinate motion control of the entire wheel-legged system. The effectiveness of the proposed control method is validated through three representative simulation scenarios in the Mujoco environment. Results demonstrate the robot's capability to maintain stable end-effector height and orientation under various conditions, including undulating terrain and flat ground. Moreover, it successfully executes multiple wheeled locomotion modes with reliable control performance. Furthermore, experiments conducted on a custom-built wheel-legged robotic platform validate the practical feasibility and effectiveness of the proposed control framework in real-world deployments.
\par
Compared with conventional wheel-legged robots employing hip-actuated mechanisms, the proposed omnidirectional design offers superior mobility and control versatility. It enables robust traversal over uneven terrain, while achieving agile omnidirectional maneuvers, and precise manipulation.
\par
Despite these advantages, hardware experiments revealed certain limitations of the proposed control framework, most notably its strong reliance on preception data and its sensitivity to modeling inaccuracies. In the future, we will further enhance the robustness and adaptability of the proposed control strategy.

\section*{Appendix} 
Let the $i$-th wheel module be fixed at $p_i\in\mathbb{R}^2$ in the body frame, with steering angle $\delta_i$, wheel spin angular velocity $\dot\phi_i$, and wheel radius $\rho$. Define,
$$e_{\parallel}(\delta_i)=\begin{bmatrix}\cos\delta_i \\ \sin\delta_i\end{bmatrix},\,\, e_{\perp}(\delta_i)=\begin{bmatrix}-\sin\delta_i\\ \cos\delta_i\end{bmatrix},\,\, J=\begin{bmatrix}0&-1\\[2pt]1&0\end{bmatrix},\,\,$$ $$\xi=\begin{bmatrix}\omega_b\\ v_b\end{bmatrix}\in\mathfrak{se}(2), \,\, u=\dot{r}=\left[ \begin{array}{c}	\dot{\delta}\\	\dot{\phi}\\\end{array} \right] \in \mathbb{R} ^8.$$
The rolling-without-slipping and no-lateral-skid constraints at each wheel can be expressed as,
$$e_{\perp}(\delta_i)^{\mathrm{T}}\left(v_b+\omega_bJp_i\right)=0,$$
$$e_{\parallel}(\delta_i)^{\mathrm{T}}\left(v_b+\omega_bJp_i\right)=\rho\dot\phi_i.$$
Stacking the four wheels yields,
$$S(\delta ,p)\,\underbrace{\left[ \begin{array}{c}	v_b\\	\omega _b\\\end{array} \right] }_{\xi}=\left[ \begin{array}{c}	0\\	\rho \,\dot{\phi}\\\end{array} \right] ,$$
$$ S(\delta ,p)=\left[ \begin{matrix}	E_{\bot}(\delta )&		d_{\bot}(p,\delta )\\ E_{\parallel}(\delta )&		d_{\parallel}(p,\delta )\\\end{matrix} \right] \in \mathbb{R} ^{8\times 3},$$

with,
\begin{equation}
\begin{aligned}
    E_{\perp}(\delta) &= \begin{bmatrix} e_{\perp}(\delta_1)^\top\\ \vdots\\ e_{\perp}(\delta_4)^\top \end{bmatrix} \in \mathbb{R}^{4\times 2}, \\
    E_{\parallel}(\delta) &= \begin{bmatrix} e_{\parallel}(\delta_1)^\top\\ \vdots\\ e_{\parallel}(\delta_4)^\top \end{bmatrix} \in \mathbb{R}^{4\times 2}, \\
    d_{\perp}(p,\delta) &= \begin{bmatrix} e_{\perp}(\delta_1)^\top J p_1\\ \vdots\\ e_{\perp}(\delta_4)^\top J p_4 \end{bmatrix} \in \mathbb{R}^{4\times 1}, \\
    d_{\parallel}(p,\delta) &= \begin{bmatrix} e_{\parallel}(\delta_1)^\top J p_1\\ \vdots\\ e_{\parallel}(\delta_4)^\top J p_4 \end{bmatrix} \in \mathbb{R}^{4\times 1}.
\end{aligned}
\end{equation}
Let $B=\mathrm{diag}(0_{4\times4},\,\rho I_4)$ so that $\begin{bmatrix}0\\ \rho\dot\phi\end{bmatrix}=Bu$. The local connection (kinematic reconstruction) is
$$\xi \;=\; S^\dagger(\delta,p)\,B\,u \;=:\; -\mathcal{A}(r)\,u,$$
where $S^\dagger$ is the Moore-Penrose pseudoinverse. If the body-frame origin is chosen at the geometric center ($\sum_{i=1}^4 p_i=0$) and the wheel locations are nondegenerate ($\sum_i\|p_i\|^2>0$), then $\mathcal{A}(r)$ has full row rank three for generic steering $\delta$.
\par
(1). The proof of full column rank of $S$. \par
Using $e_{\parallel}e_{\parallel}^{\!\top}+e_{\perp}e_{\perp}^{\!\top}=I_2$ and stacking identities, we have,
$$S^{\!\top}S= \begin{bmatrix} E_{\perp}^{\!\top}E_{\perp}+E_{\parallel}^{\!\top}E_{\parallel} & E_{\perp}^{\!\top}d_{\perp}+E_{\parallel}^{\!\top}d_{\parallel}\\ \ast & d_{\perp}^{\!\top}d_{\perp}+d_{\parallel}^{\!\top}d_{\parallel} \end{bmatrix} = \begin{bmatrix} 4I_2 & B_0\\ B_0^{\!\top} & R^2 \end{bmatrix}, \quad $$ 
$$B_0:=J\sum_{i=1}^4 p_i,\ \ R^2:=\sum_{i=1}^4\|p_i\|^2$$ 
with the geometric-center choice $\sum_i p_i=0\Rightarrow B_0=0$, we have $S^{\!\top}S=\mathrm{diag}(4I_2,\ R^2)\succ0$. Hence $S$ is full column rank 3. \par
(2). Reduce $\operatorname{rank}(\mathcal{A})$ to $\operatorname{rank}(S^{\!\top}B)$. \par
Since $S$ has full column rank, its pseudoinverse equals the left inverse $S^\dagger=(S^{\!\top}S)^{-1}S^{\!\top}$. Therefore,
$$\operatorname{rank}\mathcal{A}(r)=S^\dagger B=(S^{\!\top}S)^{-1}\,\underbrace{S^{\!\top}B}_{\text{3}\times 8}.$$
Premultiplication by the invertible $(S^{\!\top}S)^{-1}$ does not change rank, hence,
$$\operatorname{rank}\mathcal{A}(r)=\operatorname{rank}\big(S^{\!\top}B\big) \le \min \{\mathrm{rank}(S^{\!\top}),\,\,\mathrm{rank(B)}\}.$$
Obviously, $\mathrm{rank}(B)=4$, and $\mathrm{rank}(S^{\!\top})=\mathrm{rank}(S) = 3$, so that, $\mathrm{rank}\mathcal{A}(r)=3$. 

\bibliographystyle{IEEEtran}

\bibliography{IEEEabrv,ref}

\end{document}